\documentclass[a4paper,twoside, textwidth = 24cm, 10pt]{article}
\usepackage{geometry}
\usepackage{tikz}
\usepackage{tikz-3dplot}

\usepackage{pgfplots}
\usepackage{adjustbox}
\usepackage{amssymb,amsthm}
\usepackage{comment}

\usetikzlibrary{calc,patterns,angles,quotes}
\usepackage{graphicx, amsmath, color}
\usepackage{hyperref}

\usepackage{authblk}
\usepackage{xargs}
\usepackage[colorinlistoftodos,prependcaption,textsize=tiny]{todonotes}

\pgfplotsset{compat=1.17}

\newtheorem{theorem}{Theorem}[section]
\newtheorem*{theorem A}{Theorem A}
\newtheorem*{theorem B}{N\"olker's Theorem}

\newtheorem{corollary}{Corollary}[section]

\theoremstyle{remark}
\newtheorem{remark}{Remark}[section]
\theoremstyle{remark}

\theoremstyle{definition}
\newtheorem{definition}{Definition}[section]
\newtheorem{example}{Example}[section]
\numberwithin{equation}{section}

\newcommand{\vp}{\varphi}

\newcommand{\al}{\alpha}
\newcommand{\be}{\beta}

\newcommand{\Cc}{\mathcal{SC}^{\text{onf}}}

\newcommand{\Cs}{\mathrm{C}^{\text{onf} } }
\newcommand{\R}{\mathbb{R}}

\newcommand{\vth}{\vartheta}

\newcommand{\Pol}{\mathcal{P}}
\newcommand{\Q}{\mathcal{Q}}
\DeclareMathSizes{9}{9}{7}{7}

\title{On the Configurations of Closed Kinematic Chains in three-dimensional Space}
\author{Gerhard Zangerl}
\affil{Department of Mathematics, University of Innsbruck\\
Technikestra{\ss}e 13, 6020 Innsbruck, Austria\\
E-mail: {\tt gerhard.zangerl@uibk.ac.at}}
\author{Alexander Steinicke}
\affil{Department of Applied Mathematics and Information Technology,\\ Montanuniversitaet Leoben\\
Peter Tunner-Stra{\ss}e 25/I, 8700 Leoben, Austria\\
E-mail: {\tt alexander.steinicke@unileoben.ac.at}}

\date{}

\begin{document}

\maketitle

\begin{abstract}
\noindent A kinematic chain in three-dimensional Euclidean space consists of $n$ links that are connected by spherical joints. Such a chain is said to be within a closed configuration when its link lengths form a closed polygonal chain in three dimensions. We investigate the space of configurations, described in terms of joint angles of its spherical joints, that satisfy the the loop closure constraint, meaning that the kinematic chain is closed. In special cases, we can find a new set of parameters that describe the diagonal lengths (the distance of the joints from the origin) of the configuration space by a simple domain, namely a cube of dimension $n-3$. We expect that the new findings can be applied to various problems such as motion planning for closed kinematic chains or singularity analysis of their configuration spaces. To demonstrate the practical feasibility of the new method, we present numerical examples.   
\end{abstract}


\section{Introduction}

This study is the natural further development of \cite{zangerl2019configuration} in which closed configurations of a two-dimensional kinematic chain (KC) in terms of its joint angles were considered. As a generalization, we study the configuration spaces of a three-dimensional closed kinematic chain (CKC) with $n$ links in terms of the joint angles of its spherical joints. Closed configurations can be described as links that are connected by spherical joints that form a closed polygonal chain. In many applied scientific fields such as robotics, physics, computational biology, protein kinematics, and computer graphics CKCs naturally appear \cite{han2008conv}, which is why understanding their configuration spaces is extremely important.\\ 

For example, the knowledge of the configuration spaces of CKCs is important in robotics, since a robot is often required to form a closed polygonal chain while performing tasks. An example that is not so often considered appears in  physics. If $n$ force vectors acting on a point are intended to cancel each other out, this is exactly the case if they form a closed polygonal chain when they are attached to each other. A practical example of such a situation in two dimensions would be a continuous version of the balanced centrifuge problem \cite{Mat2021}. In all applications the closedness condition is typically described by nonlinear equations defining the configuration space of the CKC under consideration.\\  

The configuration space defined by those equations is a manifold \cite{JamesRTrinkleJ02a} up to points that correspond to singular positions of the CKC. If additional constraints like obstacles, link-link avoidance, or limited joint angles are considered, the configuration space is even more complicated. To deal with the complex nature of configuration spaces, three main strategies that are either based on probabilistic, algebraic or geometric methods have been developed so far. Recently, also analytic methods exploiting the Moore-Penrose pseudo-inverse were used to obtain solutions of the inverse kinematic problem for redundant CKCs \cite{tondu2015closed,ruggiu2021investigation}. We will briefly describe the three main strategies and then present our approach.\\

Algebraic methods have a long tradition in the inverse kinematics of mechanisms. They rely on the fact that the forward kinematics of a chain can be described by means of successive execution of elements of the special Euclidean group. Conversely, if one requires the KC to reach a certain endpoint with prescribed orientation, one obtains trigonometric equations that are typically algebraized by the half tangent substitution \cite{birlescu2020joint,husty2007algebraic}. Other innovative approaches use the fact that the special Euclidean motion group can be algebraically modeled by the so called study quadric. This description subsequently turns out to be useful for the inverse kinematic of mechanisms in a variety of problems.  \cite{brunnthaler2005new,hegedus2015four,husty2007new}\\

Probabilistic or randomized methods have proven to be extremely successful for motion planning in the case of situations with many practical constraints. Besides the closed-loop constraint, certain joint angles may not be possible if obstacle or link-link collisions have to be avoided. These methods are especially important for highly redundant KCs with many joints that have  high dimensional configuration spaces. Typically, when applying probabilistic methods, a random configuration is generated and then it is checked if the desired constraints are satisfied. By repeating this process, a discrete version of the configuration space can be sampled. The sampled points can then be interpolated, yielding an approximate version of the configuration space that is very useful in applications. Without further knowledge about the structure of the configuration spaces, their sampling can prove to be a difficult task. Nevertheless, probabilistic methods have been applied in different situations, which can be found in \cite{CorSim04,SLaV99,SLaV06,zhang2013unbiased,CHIRI98,CelCreRos,jaillet2013path,yakey2001randomized,suh2011tangent}.\\

The structure of configuration spaces is in itself an interesting and rich field of research and has therefore attracted the attention of many researchers. Their insight about the global geometry of configuration spaces is very important in applications and can for example be applied to improved sampling \cite{KapMil95, HAUKNU98}. In their fundamental work, Kapovitch and Milgram \cite{KapMil95} established important results about the geometry of KCs, which led to novel path planning algorithms. Further, in \cite{JamesRTrinkleJ02a,JamesRTrinkleJ02b} it is shown that if a CKC has three long links its configuration space has a particularly simple structure since in this case it is diffeomorphic to the disjoint union of two tori. Also, for the more difficult situation when CKCs do not have three long links, algorithms were derived in\cite{JamesRTrinkleJ02a,JamesRTrinkleJ02b}. Further, they developed path planners in the case of $p$ point obstacles in the plane \cite{liu2005toward,liu2020motion}. Also cohomology, homology groups and singularities of configuration spaces have been investigated in  \cite{HAUKNU98,kamiyama2007homology,blanc2012generic,TAKHIS98}. Another approach using geometric methods was developed by Han, Rudolph and Blumenthal. They discovered that it is advantageous to describe the configuration space of a CKC not by joint angles but by its diagonal lengths $L_i$, where the $L_i$ are the length of the segment connecting the origin $0$ and the $i$-th joint of the CKC, see \cite{han2008conv, han2006inverse,han2007unified}. It turns out that the lengths of possible diagonals of a CKC with $n$ links can be computed as the solution of a system of linear inequalities and as such form a convex polytope of dimension $n-3$. This polytope can be treated with methods from linear programming  \cite{luenberger2015linear}. In dimension three, feasible diagonal lengths $L_i$ correspond to infinitely many configurations of the CKC, since any configuration of a CKC can be rotated around its diagonals \cite{han2007unified}. Because of the convex structure of the space of feasible diagonals the proposed approach is very useful in applications. It has, for example, been applied to the task of motion planning very successfully in two and three dimensions, see e.g. \cite{han2006inverse, han2008conv}, where paths between CKCs with $1000$ links are computed very efficiently.\\

\textbf{Contribution:} We develop a new method that explicitly computes configurations of a CKC with $n$ in term of the joint angles of its spherical joints. The new method presented is closely related to the one in \cite{han2007unified,han2006inverse}. In our approach, the inequalities for the diagonal lengths naturally appear by manipulation of a trigonometric equation, which describes the loop closure condition for a KC. To be more precise, it turns out that this equation allows a kind of backward substitution, which is an unexpected and new mathematical insight. Another contribution is that, unlike \cite{han2007unified,han2008conv}, we investigate the linear inequalities for the diagonal lengths in our context even further. While the system of inequalities was previously handled by using linear programming methods, we are able to describe its solution set by a sampling procedure that determines the intervals of diagonals in each step by very intuitive conditions. Moreover, for the special class of CKCs that have three long links we are able to describe the diagonal lengths of such CKCs by a map that is defined on an $n-3$ dimensional cube $I_{n-3}$. A byproduct  of our explicit computations also yields a new proof for the connectedness of the configuration space for three-dimensional CKCs. Furthermore, once diagonal lengths are determined, our calculations allow to explicitly calculate the joint angles of the spherical joints of a CKC. More precisely, our method parameterizes intersecting circles of spheres, which arise  when a configuration is geometrically constructed from its diagonal lengths. Our theoretical results are confirmed by the numerical examples. We are able to calculate a configuration of a CKC with a length of one million links. To our knowledge, this has not yet been achieved in the literature so far.\\

\textbf{Outline of this text:} In section \ref{sec:sampling} we give a mathematical description of a CKC and its configuration space. Then the basic algorithm that explicitly describes how configurations of a CKC can be computed is developed in section \ref{sec:CC}. In section \ref{sec:diagspace} we describe the set of new parameters and show how they can be used to compute a vector of joint angles of  a CKC. Then in section \ref{sec:diagspace} a special class of CKCs is considered, for which the diagonal space can be explicitly parameterized.   
Finally, we present numerical examples that show the validity of the developed results.

\section{Configuration space}\label{sec:sampling}

For a CKC with link lengths $a_1, \dots, a_n$ we introduce Cartesian coordinates in three-dimensional Euclidean space. Moreover, we place one of the links of the CKC so that it is lying on the positive $x$-axis and that one of its ends coincides with the origin. Without loss of generality we can assume that the link $a_n$ of the chain is fixed in the described manner, see Figure \ref{fig:ClosedChain3D}. For $1 \leq k \leq n$ denote by $ \vp^k := (\vp_1, \dots, \vp_{k})\in [0,2\pi)^k  $  and    $ \vth^k := (\vth_1, \dots, \vth_k  ) \in [0,\pi]^k$  vectors of angles and by  $a^n := (a_1, \dots, a_n) \in \R^n$  a vector of link lengths. We set $M^k := [0,2\pi)^{k} \times [0,\pi]^{k}$ and denote by 
\begin{align}
f_{a^n,k} \colon  M^k \rightarrow \R^3, \quad  f_{a^n ,k} \left(\vp^k, \vth^k \right) =    \sum_{j=1}^{k}  a_j 
\begin{pmatrix} 
\sin(\vth_j) \cos(\vp_j)  \\
\sin(\vth_j) \sin(\vp_j)  \\ 
\cos(\vth_j)
\end{pmatrix}.
\end{align}
the $k$-th endpoint map that attaches the first $k$ links of the KC with prescribed direction together. We will refer to the set of locations $f_{a^n,k}\left( M^k \right)$ that can be reached by it as its workspace. We will call $(\vp^{n-1}, \vth^{n-1} )$ a configuration of the CKC with link lengths $a_1, \dots,a_n$ if it satisfies the (loop) closure condition, which means that it is contained in the set 
\begin{align}\label{clos_eq}
&& \Cs(a^n) = \left\{   (\vp^{n-1}, \vth^{n-1} )  \in M^{n-1}  \colon  f_{a^n,n-1}\left( \vp^{n-1} , \vth^{n-1}  \right) =
\begin{pmatrix} a_n \\ 0 \\ 0 
\end{pmatrix} \right\} = f_{a^n, n-1}^{-1} \left(a_n, 0 , 0 \right).
\end{align}
 If no restrictions on the endpoint map are imposed, $\left( \vp^{n-1} , \vth^{n-1}\right) \in M^{n-1}$ will just be called a configuration of the kinematic chain (KC) with $n-1$ links. Note that in our case the configuration space of a CKC is described in terms of its absolute joint angles. Other definitions describe configurations of a CKC as a real algebraic variety given by   
\begin{align*}
    \Cs_{\text{alg}}(a^n) := \left\{ p = (p_1, \dots , p_n ) \in \left( \R^3\right)^n \colon p_1 = 0,~ p_n=(a_n,0,0)^T,~ ||p_j-p_{j+1}|| = a_j, ~  1 \leq j \leq n-1      \right\}. 
\end{align*}
Here $\| \cdot  \|_2 $ denotes the Euclidean norm. We point out that this two definitions are equivalent, since a configuration described by joint angles obviously can be used to compute a $p \in \Cs_{\text{alg}}(a^n)$ and vice versa. 
\begin{figure}
\begin{center}
\tdplotsetmaincoords{60}{110}
\pgfmathsetmacro{\rvec}{1.2}
\pgfmathsetmacro{\thetavec}{-30}
\pgfmathsetmacro{\phivec}{60}
\begin{tikzpicture}[scale=1.5,tdplot_main_coords]
\coordinate (O) at (0,0,0);
\draw[->] (0,0,0) -- (2,0,0) node[anchor=north east]{$x$};
\draw[->] (0,0,0) -- (0,2,0) node[anchor=north west]{$y$};
\draw[->] (0,0,0) -- (0,0,2) node[anchor=south]{$z$};
\tdplotsetcoord{P}{\rvec}{\thetavec}{\phivec}
\draw[color=black,thick] (O) -- (P);
\draw[color=black,thick] (P) -- (1.5,2,2.5);
\draw[color=black,thick]  (1.5,2,2.5)--(3,1,2.5);
\draw[color=black,thick]  (3,1,2.5)--(1.5,0,0);
\shade[ball color=red!30!white] (0,0,0) circle (1.5pt);
\shade[ball color=red!30!white] (P) circle (1.5pt);
\shade[ball color=red!30!white] (1.5,2,2.5) circle (1.5pt);
\shade[ball color=red!30!white] (3,1,2.5) circle (1.5pt);
\shade[ball color=red!30!white] (1.5,0,0) circle (1.5pt);
\fill[ball color=blue, opacity=0.15] (0,0,0) circle (0.88cm);
\node at (0.75, 0.2, 0 ) {$a_5$}; 
\end{tikzpicture}
\hspace{0.5cm}
\begin{tikzpicture}[scale = 0.9]
\begin{axis}[ 
axis x line  = middle, 
axis y line = middle, xmin = 0, xmax = 5, ymin = -0, ymax = 5, 
xtick = \empty,  
ytick = \empty,
grid = major, enlargelimits = 0.2]
\draw (axis cs: 0, 0)--(axis cs: 1, 3);
\draw (axis cs: 2.5, 3)-- (axis cs:4, 2);
\draw (axis cs: 4, 2)--(axis cs: 5, 0); 
\draw (axis cs: 0, 0)--(axis cs: 4, 2);
\draw (axis cs: 0, 0)--(axis cs: 2.5, 3);
\draw[dashed] (axis cs: 1, 3)--(axis cs: 2.5, 3);
\draw[fill = red] (axis cs: 0 ,0 )   circle (3pt);
\draw[fill = red] (axis cs: 1 ,3 )   circle (3pt);
\draw[fill = red] (axis cs: 4 ,2 )   circle (3pt);
\draw[fill = red] (axis cs: 5 ,0 )   circle (3pt);
\draw[fill = red] (axis cs: 2.5 ,3) circle (3pt);
\node at (axis cs:2.5, -0.5 ) {\fontsize{12}{12}$a_n = L_{n-1}\left(\alpha^{n-1}, \beta^{n-1}  \right)$};
\node at (axis cs:3.5, 2.8 ) {\fontsize{12}{12} \rotatebox{-25}{$a_{n-2}$}};
\node at (axis cs:4.8, 1 ) {\fontsize{12}{12} \rotatebox{-60}{$a_{n-1}$}};
\node at (axis cs:0.3, 2 ) {\fontsize{12}{12} \rotatebox{72}{$a_{1} = L_1 $}};
\node at (axis cs:1.95, 1.3 ) {\fontsize{12}{12} \rotatebox{25} { \fontsize{10}{10} $L_{n-2}\left( \alpha^{n-2},\beta^{n-2} \right) $}};
\end{axis}
\end{tikzpicture}\caption{Left: A three-dimensional CKC with $n=5$ five links. Spherical joints are illustrated as small spheres. The first joint of the chain is the origin and link $a_5$ is supported on the positive $x$-axis. The transparent ball indicates all the possible locations of the endpoint of spherical configurations. Right: The  diagonal lengths of CKC are depicted. Since their definition is the same for any dimension, for simplicity a configuration of a CKC in two dimensions is depicted.}\label{fig:ClosedChain3D} 
\end{center}
\end{figure}
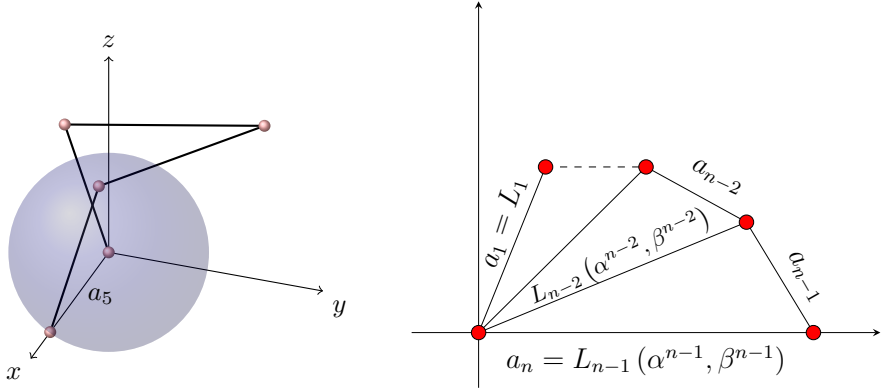\\

The analysis carried out in in this work uses the simple observation that it is sufficient to understand the space \begin{align}\label{eq:circular_configurations}
\Cc(a^n)   =  \left\{  \left(    \alpha^{n-1}, \beta^{n-1} \right) \in M^{n-1}  \colon   \| f_{a^n,n-1}\left(\alpha^{n-1} ,\beta^{n-1}	\right) \|^2_2  = a_n^2 \right\},   
\end{align}
in order to describe $\Cs(a^n)$. From the definition of $\Cc(a^n) $ it is clear that for any configuration $ \left(  \alpha^{n-1} , \beta^{n-1}  \right) \in \Cc(a^n)$ its endpoint satisfies  
\begin{alignat*}{2}
f_{a^n ,n-1}\left(  \alpha^{n-1} , \beta^{n-1}  \right) \in S_{a_n}, 
\end{alignat*}
where $S_{a_n}$  is the sphere that is centred at the origin and has radius $a_n$.  We will refer to such a configuration $\left(  \alpha^{n-1} , \beta^{n-1}  \right)$ as \textit{spherical configuration} of a CKC. Suppose that $ f_{a, n-1}\left(  \alpha^{n-1} , \beta^{n-1}  \right) $  is the endpoint  of a spherical configuration  $\left(  \alpha^{n-1} , \beta^{n-1}  \right)$ and has spherical coordinates $(\lambda, \mu)$ on $S_{a_n}$. Then there is a rotation  $R_{\lambda, \mu}$ such that,   
\begin{alignat*}{2}
R_{\lambda, \mu} \cdot  f_{a^n, n-1}\left(  \alpha^{n-1} , \beta^{n-1}  \right) = 
\begin{pmatrix}
a_n\\0\\0
\end{pmatrix}. 
\end{alignat*} 
The latter expression can be used to obtain $(\vp^{n-1}, \vth^{n-1} ) \in \Cs_{a}$ from a spherical configuration. This argumentation shows that an efficient method of calculating the solutions of the implicit equation 
\begin{alignat}{2}\label{eq:closed_norm}
L_{n-1}^2 (\alpha^{n-1},\beta^{n-1}) :=  \| f_{a^n,n-1}\left( \alpha^{n-1} , \beta^{n-1}	\right) \|^2_2  = a_n^2, 
\end{alignat}
also leads to an efficient method to determine closed configurations from $\Cs(a^n)$. Here $L_{n-1}$ is a diagonal of the CKC, see Figure \ref{fig:ClosedChain3D} and also section \ref{sec:Toolnot}.\\

Summarizing, we  obtain such configurations of a CKC by the following two step algorithm:
\begin{itemize}
\item[(i)]   Compute a spherical configuration  $ \left( \alpha^{n-1} , \beta^{n-1} \right) \in \Cc(a^n)$\\
\item[(ii)]  Determine $  (\lambda , \mu)$ and $R_{\lambda, \mu}$ and compute   the closed configuration $\left( \vp , \vth \right)$.  
\end{itemize}
Since step (ii) is a rather easy task once a spherical configuration is obtained we will focus on step (i). In this step we are concerned with describing the solutions of the trigonometric equation \eqref{eq:closed_norm}, which is in its expanded form given by 
\begin{align}
\left(  \sum_{j = 1}^{n-1} a_j \sin(\be_j) \cos(\al_j)   \right)^2  +  \left(  \sum_{j = 1}^{n-1} a_j \sin(\be_j) \sin(\al_j)   \right)^2 + \left(  \sum_{j = 1}^{n-1} a_j \cos(\be_j)   \right)^2 = a_n^2.   
\end{align}
Rearranging and simplifying according to trigonometric addition formulas gives 
\begin{align}\label{eq:CKCsquared_expanded}
\sum_{i=1}^{n-1} a_i^2 + 2 \sum_{i<j}^{n-1} a_i a_j \sin\left( \be_i  \right) \sin\left( \be_j  \right) \cos(\al_i -\al_j)    + 2 \sum_{i< j}^{n-1} a_i a_j \cos(\be_i)  \cos(\be_j)  
= a_n^2 . 
\end{align}
Consider  the map  $g \colon M^{n-1} \rightarrow \R$ with   $g \left(\al^{n-1}, \be^{n-1} \right) := \| f_{a^n, n-1} \left(\al^{n-1}, \be^{n-1} \right)   \|^2_2$. Then $ \Cc_a = g^{-1}(a_n^2)$ and by the preimage theorem we know that  the set $\Cc_a$ of all spherical configurations is a sub-manifold of $M^{n-1}$ of dimension $2n-3$, whenever $a_n^2$ is a regular value of the map $g$. In all other cases the space $\Cc_{a}$ may have singular points. Although the geometry of this space might be complicated it turns out solutions to the trigonometric equation  \ref{eq:CKCsquared_expanded} can be obtained by a kind of backwards substitution. This computations are outlined in detail in section \ref{sec:CC} after we introduce notations and state necessary mathematical tools to do so in section \ref{sec:Toolnot}.  

\subsection{Mathematical tools and notations}\label{sec:Toolnot}

\begin{figure}
\centering 
\begin{tikzpicture}[scale = 0.8]
\begin{axis}[ 
axis x line  = middle, 
axis y line = middle, xmin = -2, xmax = 2, ymin = -0, ymax = 2, 
xtick = \empty,  
ytick = \empty,
grid = major, enlargelimits = 0.2]
\draw[->, -latex, line width = 1pt] (axis cs: 0,0)--(axis cs:-2,2);
\draw[->, -latex] (axis cs:1.4,0) arc (0:114:2cm);
\node at (axis cs:-1.8, 2.2) {\fontsize{15}{15} $P(a|b)$ };
\node at (axis cs: 0.35,0.5)  { \colorbox{white}{\fontsize{10}{10} $\arg(a,b)$}};
\node at (axis cs:2.4, -0.2) {\fontsize{15}{15} $x$};
\node at (axis cs:0.3, 2) {\fontsize{15}{15} $y$};
\draw[dashed] (axis cs: -2,0)--(axis cs: -2,2);
\draw[dashed] (axis cs:0,2)--(axis cs: -2,2);
\draw[thin, fill = red] (axis cs: 0,0) circle (3pt);
\end{axis}
\end{tikzpicture}
\caption{The function $\arg$ gives the oriented angle between the $x$-axis and the vector from the origin to $P(a|b)$.}\label{fig:Atan2}
\end{figure}
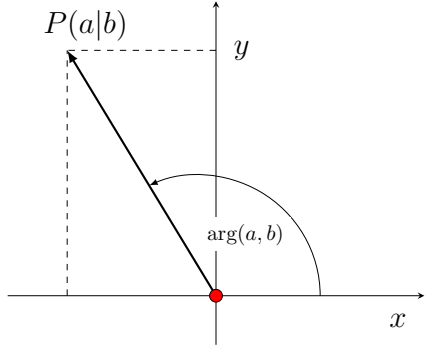

Surprisingly, the trigonometric equation \eqref{eq:CKCsquared_expanded} can be rearranged into an equation of the same type but with one joint less appearing on its left hand side, which will be done in Section \ref{sec:CC}. For the necessary computations to show this fact, we will use the addition theorem for a linear combination of sine and cosine functions: \begin{align}\label{eq:addsincos}
a \sin\left( x \right) + b \cos\left( x\right) = c \sin\left(x + \arg(a,b)  \right), 
\end{align}
where $c = \sqrt{a^2 + b^2}$.  \text{and}  $\arg \left(a,b\right) $
is the function described in Figure \ref{fig:Atan2}.
In order to achieve a compact presentation of the results that will follow it is important to introduce several abbreviations and notations. These abbreviations are motivated by an equivalent form of equation \eqref{eq:CKCsquared_expanded} that is obtained by taking the summand of index $n-1$ out of the sum, applying trigonometric addition formulas and rearranging the remaining terms. The resulting form is given by   
\begin{align}\label{eq:squaredrearrangedA}
&\cos(\al_{n-1}) 2\sum_{j = 1 }^{n-2}    a_{n-1} a_j  \sin(\be_j) \sin(\be_{n-1}) \cos(\al_j)   + \sin(\al_{n-1})  2\sum_{j=1 }^{n-2}    a_{n-1} a_j  \sin(\be_j) \sin(\be_{n-1}) \sin(\al_j)+ \nonumber  \\  & 2\sum_{i< j}^{n-2} a_i a_j \sin(\be_j) \sin(\be_i) \cos(\al_i -\al_j) + a_{n-1} \cos(\be_{n-1}) 2\sum_{j=1}^{n-2} a_j \cos(\be_j) + 2\sum_{i< j}^{n-2} a_i a_j \cos(\be_i) \cos(\be_j) +S_{n-1}= a_n^2,  
\end{align}
where $S_{n-1}: = \sum_{i=1}^{n-1} a_i^2$ and accordingly   $S_{n-k}: = \sum_{i=1}^{n-k} a_i^2$ for $1 \leq k \leq n-2$.\\

For the link lengths $a^n$ and $\left( \al^{n-1} ,\be^{n-1} \right) \in M^{n-1}$ we define for $1 \leq  k \leq  n-1 $  the following abbreviations: First denote by $X_{n-k}\left(\al^{n-k} ,  \be^{n-k} \right),Y_{n-k}\left(\al^{n-k} ,  \be^{n-k} \right)$ and $Z_{n-k}\left(\al^{n-k} ,  \be^{n-k} \right)$  the $x$-, $y$- and $z$-component of $f_{a^n, n-k} \left(\al^{n-k} ,  \be^{n-k} \right)$, that is, 

\begin{align*}
X_{n-k} \left(\al^{n-k} ,\be^{n-k}\right)	&:=  \sum_{j = 1 }^{n-k}    a_j  \sin(\be_j) \cos(\al_j),\\ 
Y_{n-k} \left(\al^{n-k}, \be^{n-k}\right) &:= \sum_{j = 1 }^{n-k}    a_j  \sin(\be_j) \sin(\al_j)\\
Z_{n-k} \left(\al^{n-k}, \be^{n-k}\right) &:= \sum_{j = 1 }^{n-k}    a_j  \cos(\be_j).
\end{align*}

Furthermore, for $2 \leq k \leq n-2$, we will refer to the values 
\begin{align}\label{def:diagsconf}
L_{n-k}\left(\al^{n-k} ,  \be^{n-k} \right) :=\left\|f_{a^n, n-k} \left(\al^{n-k} ,  \be^{n-k} \right)\right\|
\end{align}
as the {\it length of the diagonals}, of a configuration $\left( \al^{n-1},\be^{n-1} \right)$ and denote by  $L^{n-3} := (L_2, \cdots, L_{n-2})$ the vector of (variable) diagonal lengths of this configuration. We point out here that $L_1 = a_1$ and $L_{n-1} = a_n$ are fixed and are therefore not contained within $L^{n-3}$. Note that with this  framework we have that  
\begin{align*}
L_{n-k}^2\left(\al^{n-k} ,  \be^{n-k} \right) &=  X_{n-k}\left(\al^{n-k} ,  \be^{n-k} \right)^2+Y_{n-k}\left(\al^{n-k} ,  \be^{n-k} \right)^2+Z_{n-k}\left(\al^{n-k} ,  \be^{n-k} \right)^2\\
&= 2\sum_{i< j}^{n-k} a_i a_j \sin(\be_j) \sin(\be_i) \cos(\al_i -\al_j)+ 2\sum_{i< j}^{n-k} a_i a_j \cos(\be_i) \cos(\be_j)+S_{n-k},
\end{align*}
where the second equality follows from the addition formula for the cosine function the same way as \eqref{eq:CKCsquared_expanded} was found. Finally, we denote by
\begin{alignat}{2}
\Phi_{n-k} \left(\al^{n-k} ,\be^{n-k}\right)	&:= \arg \Biggl( \underbrace{\sum_{j=1}^{n-k} a_j \sin\left(\be_j \right) \sin(\al_j)}_{Y_{n-k} \left(\al^{n-k}, \be^{n-k}\right)} , \underbrace{\sum_{j=1}^{n-k} a_j \sin\left(\be_j \right) \cos(\al_j)}_{X_{n-k} \left(\al^{n-k}, \be^{n-k}\right)}   \Biggr) \\ 
\Psi_{n-k} \left(\al^{n-k}, \be^{n-k-1}\right) &:=  
\arg \Biggl( \sin \left(  \al_{n-k} + \Phi_{n-k-1}\left( \al^{n-k-1}, \be^{n-k-1}\right)\right)C\left(\al^{n-k-1}, \be^{n-k-1}\right),\!\!\!\!\underbrace{\sum_{j = 1}^{n-k-1} a_j\cos(\be_j)}_{Z_{n-k-1}\left(\al^{n-k-1} ,  \be^{n-k-1} \right)}    \!\!\!    \Biggr),\\
&\quad \text{with }C\left(\al^{n-k-1}, \be^{n-k-1}\right):=\sqrt{X_{n-k-1} \left(\al^{n-k-1}, \be^{n-k-1}\right)^2+Y_{n-k-1} \left(\al^{n-k-1}, \be^{n-k-1}\right)^2}
\label{eq:Psi}
\end{alignat}
In calculations we will often omit the arguments $\left(\al^{n-k-1}, \be^{n-k-1}\right)$ for readability. We point out that in its second argument, $\Phi_{n-k}$ depends on the $\beta$-angle components up to $\be_{n-k}$ whereas $\Psi_{n-k}$ only depends on those up to $\be_{n-k-1}$.

\section{Sampling the space of spherical configurations}\label{sec:CC}

Spherical configurations are implicitly given as solutions of the trigonometric equation \eqref{eq:CKCsquared_expanded}. Although such an equation is useful to determine whether a configuration $(\alpha^{n-1},\beta^{n-1} )$ satisfies the closedness condition or not, a more explicit description of the configuration space is highly desirable. It turns out that the desired description can be obtained by manipulating equation \eqref{eq:CKCsquared_expanded}. This results in an algorithm to systematically determine solutions to \eqref{eq:CKCsquared_expanded}, which is outlined in Corollary \ref{thm:sampling}. For the reader's better comprehensibility we included a preparatory section. 

\subsection{Mathematical prerequisites}\label{ssec:prereq}

Equation \eqref{eq:CKCsquared_expanded} can be compactly written as $L_{n-1}^2=a_n^2$. Since our goal is to manipulate this equation, we write it as     
\begin{align}
2\sum_{i<j}^{n-1} a_i a_j \sin\left( \be_i  \right) \sin\left( \be_j  \right) \cos(\al_i -\al_j)    +  2\sum_{i< j}^{n-1} a_i a_j \cos(\be_i)  \cos(\be_j) +S_{n-1} 
&= a_n^2\nonumber .
\end{align}
Our reduction step of the equation $L_{n-1}^2 = a_n^2$ is now based on a trick that brings the diagonal length $L_{n-2}$ into play. This can be achieved by fixing the element with index $n-1$ in the sums and applying the addition formula for the cosine function to $\cos(\alpha_{i}-\alpha_{n-1})$ and using the notation from the last section to rewrite \eqref{eq:squaredrearrangedA} as 
\begin{align}\label{eq:squaredrearrangedB}
\cos(\al_{n-1}) \sin(\be_{n-1})a_{n-1}2 { X_{n-2} }   &+ \sin(\al_{n-1})  \sin(\be_{n-1}) a_{n-1}2 {Y_{n-2}} \nonumber\\
&+a_{n-1} \cos(\be_{n-1}) 2 {Z_{n-2}}  + {L_{n-2}^2} = L_{n-1}^2-a_{n-1}^2,    
\end{align}
where we used that $L_{n-1} = a_n$. The original equation \eqref{eq:closed_norm} is equivalent to the latter equation. However, in this transformed version of the equation the diagonal length of the mechanism reduced by one link $L_{n-2}$ appears. Moreover, one recognizes that it is already tempting to apply \eqref{eq:addsincos} for the angle $\al_{n-1}$. This observation will be the basis of Theorem \ref{thm:reduction}. 

\subsection{Computation and sampling of spherical configurations}\label{sec:CCc} 

In order to achieve a description of the configuration space, we show that trigonometric equation \eqref{eq:CKCsquared_expanded} allows for a kind of backward substitution, see Theorem \ref{thm:reduction}. Subsequently, this property can then be exploited to determine configurations by a successive procedure. The approach is similar to the one in  \cite{zangerl2019configuration}, but the presentation is more mature and clearer in this follow-up work. We introduce a definition before we state the main result of the chapter.

\begin{definition}[Diagonal Space]\label{def:DiagSpace} For a KC with link lengths $a_1,\dots, a_n$, recall that $L^{n-3} \in \mathbb{R}^{n-3}_{\geq 0}$. We denote the diagonal space of the KC as the set \begin{align}
&\mathcal{DS}({a^n}):= \left\{  L^{n-3}  \colon L_{n-k-1} \in   
\Bigl[\left|L_{n-k}-a_{n-k}\right|,L_{n-k}+a_{n-k}\Bigr] \cap  
\Bigl[0\vee \mathrm{    R}_{n-k-1}^{\min}, \mathrm{    R}_{n-k-1}^{\max}\Bigr], ~1\leq k \leq n-3 \right\},   
\end{align}
where we use the notation 
\begin{align}\label{eq:cmincmaxA}
\mathrm{    R}_{k}^{\min}   :=  \max_{1\leq i\leq k}\left(2a_i-\sum_{j=1}^ka_j\right) ~ \text{and}~ \mathrm{    R}_{k}^{\max} :=    \sum_{j=1}^ka_j  ~~\text{for}~1\leq i\leq k.
\end{align}
Moreover we use the abbreviation $0 \vee  \mathrm{    R}_{k}^{\min} := \max\left\{ 0,  \mathrm{    R}_{k}^{\min}  \right\}$.  
\end{definition}

Note that the quantities $\mathrm{R}_{k}^{min}$ and $\mathrm{R}_{k}^{max}$ also play an important role in \cite{JamesRTrinkleJ02b}. We state our main result about CKCs.   

\begin{theorem}[Chain reduction]\label{thm:reduction} Suppose  $\left( \alpha^{n-1}, \beta^{n-1}\right) \in M^{n-1}$ is a configuration of a KC for a given vector of link lengths $a^n$. Then $\left( \alpha^{n-1}, \beta^{n-1}\right) \in \Cc(a^n)$ if and only if its vector of its diagonals $L^{n-3}=(L_2,\dotsc,L_{n-2})$ is an element of $\mathcal{DS}({a^n})$. Furthermore, the angles $\al_{n-k}, \be_{n-k}$ are related to the remaining vector of angles $\al^{n-k-1}, \be^{n-k-1}$ of the spherical configuration, for  $1\leq k\leq n-2$, by the equation (that is analogue to \eqref{eq:squaredrearrangedB}) 
\begin{align}\label{eq:CGeqA0}
\cos(\al_{n-k}) \sin(\be_{n-1})a_{n-k}2 { X_{n-k-1} }   &+ \sin(\al_{n-k})  \sin(\be_{n-k}) a_{n-k}2 {Y_{n-k-1}} \nonumber \\ &+a_{n-k} \cos(\be_{n-k}) 2 {Z_{n-k-1}}  + {L_{n-k-1}^2} =  L_{n-k}^2-a_{n-k}^2,
\end{align}
which is equivalent to
\begin{align}\label{eq:CGeqA}
2a_{n-k}  \sin \left(\be_{n-k} + \Psi_{n-k}  \right) \left[  \sin \left(\al_{n-k} +\Phi_{n-k-1} \right)^2  \left(   X_{n-k-1}^2  + Y_{n-k-1}^2 \right) + Z_{n-k-1}^2   \right]^{\frac{1}{2}}  
&+ L_{n-k-1}^2 \nonumber\\  
&=  L_{n-k}^2  -a_{n-k}^2, 
\end{align}
whenever all expressions therein are defined. The latter equality explains the name of the theorem, since it explicitly relates sub-chains with $n-k$ and $n-k-1$ links of the CKC.   
\end{theorem}

\begin{proof} Assume a spherical configuration $(\al^{n-1}, \be^{n-1})$ is given. We will manipulate equation \eqref{eq:squaredrearrangedB} to show that the diagonal lengths are indeed in $\mathcal{DS}({a^n})$. We show the result for the diagonal $L_{n-2}$. For the remaining diagonals, the argument then follows inductively by repeatedly applying the same arguments.\\   

First consider the case when $X_{n-2} = Y_{n-2} = 0$ and $Z_{n-2} \neq 0$. Then the value of $\al_{n-1}$ does not play a role in \eqref{eq:CGeqA0}. (That is, if conversely a vector of diagonals $L^{n-3}$ is given, $\al
_{n-1}$ can be chosen arbitrarily in $[0,2 \pi]$.) We see that $ |Z_{n-2}| = L_{n-2}$ holds and get that 
\begin{align*}
2a_{n-1}\cos(\beta_{n-1})Z_{n-2} + Z_{n-2}^2 = L_{n-1}^2-a_{n-1}^2. 
\end{align*}
The latter equation can be solved for $\be_{n-1}\in [0,\pi]$ iff  the inequalities  
\begin{align*}
-a_{n-1} \leq \frac{ L_{n-1}^2-a_{n-1}^2 - L_{n-2}^2 }{2 Z_{n-2}} \leq a_{n-1},
\end{align*}
are satisfied, or equivalently if the diagonal length  satisfy the inequality $4 a_{n-1}^2 L_{n-2}^2 \geq \left( L_{n-1}^2-a_{n-1}^2 - L_{n-2}^2 \right)^2$. showing that $L_{n-2}$ satisfies  $L_{n-2} \in \Bigl[ |L_{n-1} - a_{n-1}| , L_{n-1} + a_{n-1}    \Bigr]$. Furthermore, since $(\al^{n-1},\be^{n-1})$ is a spherical configuration the diagonal $L_{n-2}$ must also be reachable by the remaining links of the CKC, see \ref{rem:splitting}. Therefore  $L_{n-2} \in \Bigl[ \mathrm{R}_{n-1}^{min} , \mathrm{R}_{n-1}^{max} \Bigr]$  has to be satisfied and in this case we can  conclude that  
\begin{align}\label{set:DSiterative}
L_{n-2} \in \Bigl[ |L_{n-1} - a_{n-1}| , L_{n-1} + a_{n-1}    \Bigr] \cap  \Bigl[ \mathrm{R}^{\min}_{n-1} , \mathrm{R}^{\max}_{n-1}\Bigr].  
\end{align}
We mention that also in the trivial case when $ X_{n-2}= Y_{n-2} = Z_{n-2} = 0$, this condition is trivially satisfied and $L_{n-1}= a_n = a_{n-1}$ must hold.\\ 

After having dealt with the special cases, we can assume that $(X_{n-2}, Y_{n-2})\neq (0, 0)$. Then applying formula \eqref{eq:addsincos} to equation \eqref{eq:squaredrearrangedB}  gives
\begin{align}\label{eq:addthmA}
&2a_{n-1} \sin \left(\al_{n-1} +\Phi_{n-2} \right) \sin(\be_{n-1}) \left[ X_{n-2}^2 + Y_{n-2}^2   \right]^{1/2}   
+ 2a_{n-1} \cos(\be_{n-1}) Z_{n-2} + L_{n-2}^2 = L_{n-1}^2-a_{n-1}^2.
\end{align}
Note that we can apply formula \eqref{eq:addsincos} again to equation \eqref{eq:addthmA} for $\beta_{n-1}$ and obtain 
\begin{align}\label{eq:addthmtwice}
2a_{n-1}  \sin \left(\be_{n-1} + \Psi_{n-1} \right) \left[  \sin \left(\al_{n-1} +\Phi_{n-2} \right)^2  \left(   X_{n-2}^2  + Y_{n-2}^2 \right) + Z_{n-2}^2   \right]^{\frac{1}{2}}  
+  L_{n-2}^2  = L_{n-1}^2-a_{n-1}^2  
\end{align}
provided that the expression in the square brackets is not zero. If so, $Z_{n-2}  $ must vanish and we conclude that $\sin(\alpha_{n-1} + \Phi_{n-2}) = 0$ and therefore $\al_{n-1} = 2\pi -\Phi_{n-2}$ or $\al_{n-1} = 2\pi -\Phi_{n-2}$, whereas regardless of the choice for $\al_{n-1}$ the angle $\be_{n-1}$ can be chosen arbitrarily in $[0,\pi]$. Note that also in these cases \eqref{set:DSiterative} is satisfied. 
Finally, as  $\sin \left(\al_{n-1} +\Phi_{n-2} \right)^2  \left(   X_{n-2}^2  + Y_{n-2}^2 \right) + Z_{n-2}^2 \neq 0$, solving \eqref{eq:addthmtwice} for $\beta_{n-1}$ is only possible if and only if the chain of inequalities 
\begin{align*}
-a_{n-1} \leq \frac{ L_{n-1}^2-a_{n-1}^2 -  L_{n-2}^2   }{    
2\left[  \sin \left(\al_{n-1} +\Phi_{n-2} \right)^2  \left(   X_{n-2}^2  + Y_{n-2}^2 \right)    +   Z_{n-2}^2   \right]^{\frac{1}{2}}  
} \leq a_{n-1}
\end{align*}
are satisfied and lead immediately to the equivalent inequality  
\begin{align}\label{eq:CGineq}
\left(  L_{n-1}^2-a_{n-1}^2 -  L_{n-2}^2  \right)^2 \leq 4 a_{n-1}^2 \bigg(\sin \left(\al_{n-1} +\Phi_{n-2} \right)^2   \left(   X_{n-2}^2 +  Y_{n-2} ^2       \right) +  Z_{n-2}^2\bigg).
\end{align}
Further rearranging gives 
\begin{align}\label{eq:CGineqsine}
4a_{n-1}^2 \sin \left(\al_{n-1} +\Phi_{n-2} \right)^2 \geq  \frac{ \left(  L_{n-1}^2-a_{n-1}^2 -  L_{n-2}^2  \right)^2 -4a_{n-1}^2   Z_{n-2}^2  }{ X_{n-2}^2 + Y_{n-2}^2}.
\end{align}
If $\alpha_{n-1}$ is given, a fortiori, it follows that
\begin{align}\label{eq:CGafortiori}
4a_{n-1}^2  \geq  \frac{ \left(  L_{n-1}^2-a_{n-1}^2 -  L_{n-2}^2 \right)^2 -4a_{n-1}^2   Z_{n-2}^2  }{ X_{n-2}^2 + Y_{n-2}^2}.
\end{align} 
Since $X_{n-2}^2 + Y_{n-2}^2+Z_{n-2}^2=L_{n-2}^2$, this gives 
\begin{align}\label{eq:CGreduced}
\left(   L_{n-1}^2-a_{n-1}^2 -  L_{n-2}^2  \right)^2 \leq  4a_{n-1}^2  L_{n-2}^2, 
\end{align}
which means that  \eqref{set:DSiterative} holds for $L_{n-2}$ and hence the diagonal length is feasible. Interpreting the original equation \eqref{eq:CKCsquared_expanded} as $L_{n-1}=a_n$, we can proceed in an iterative manner by solving an equation of the form $L_{n-2} = a$, where $a$ is an element in the interval \eqref{set:DSiterative}. This  shows that a spherical configuration yields a vector of lengths $L^{n-3} \in \mathcal{DS}({a^n})$ and the relation \eqref{eq:CGeqA} holds. Conversely, assume that $ L^{n-3} \in \mathcal{DS}({a^n})$ is given. By simple equivalency transformations of the conditions in the definition of $\mathcal{DS}({a^n})$, \eqref{eq:CGreduced} follows. So does \eqref{eq:CGafortiori} and this inequality yields a range for the angle $\alpha_{n-1}$ such that \eqref{eq:CGineqsine} holds. From there, going upwards by equivalency transformations assuming all terms exist, we are guaranteed that \eqref{eq:CGeqA} is solvable and we can use it to compute a spherical configuration $(\al^{n-1}, \be^{n-1})$. The special cases, where not all terms are well-defined are similar but easier to treat.
\end{proof}

It is interesting to consider some special cases to reflect on the results of the last theorem.\\ 

\begin{figure}
\centering
\begin{tikzpicture}[scale = 0.6]
\draw[->, -latex, line width = 0.5pt] (-1,0) -- (6,0);
\draw[->, -latex, line width = 0.5pt] (0,-1) -- (0,4);
\draw[line width = 1pt] (0,0) -- ++(160:4);
\draw[line width = 1pt, dashed]  (160:4) -- ++(100:3)  ;
\draw[line width = 1pt]   (160:4)  ++(100:3) -- ++(60:3); 
\draw[line width = 1pt]   (160:4)  ++(100:3) ++(60:3) -- ++(20:2); 
\draw[line width = 1pt]  (0,0)--(3.1,0) ;
\draw[line width = 1pt]  (3.1,0) -- ++(60:5.8); 
\draw[line width = 1pt,  ]   (0,0) -- (97:7.7); 
\draw[line width = 1pt,  ]   (0,0) -- (112:7.5); 
\draw[line width = 1pt,  ]   (0,0) -- (135:6); 
\draw[line width = 1pt, ]   (0,0) -- (40:7.7); 
\draw[line width = 1pt, fill  = red, thick  ]   (0,0) circle  (5pt); 
\draw[line width = 1pt, fill  = red, thick  ]   (160:4) circle  (5pt); 
\draw[line width = 1pt, fill  = red, thick  ]   (160:4)  ++(100:3) circle  (5pt); 
\draw[line width = 1pt, fill  = red, thick  ]   (160:4)  ++(100:3) ++(60:3) circle  (5pt); 
\draw[line width = 1pt, fill  = red, thick  ]   (160:4)  ++(100:3) ++(60:3)++(20:2) circle  (5pt); 
\draw[line width = 1pt, fill  = red, thick  ]   (3.1,0)  ++(60:5.8) circle  (5pt); 
\draw[line width = 1pt, fill  = red, thick  ]   (3.1,0)  circle  (5pt); 
\node at (-2, 8 ) {\fontsize{12}{12} \rotatebox{23}{$a_{n-2}$}};
\node at (-4, 6 ) {\fontsize{12}{12} \rotatebox{60}{$a_{n-3}$}};
\node at (4.5, 1.5 ) {\fontsize{12}{12} \rotatebox{73}{$a_{n-1}$}};
\node at (1.5,-0.5 ) {\fontsize{12}{12} \rotatebox{0}{$a_{n}$}};
\node at (-2, 0 ) {\fontsize{12}{12} \rotatebox{-10}{$a_{1}  $}};
\node at (2, 2.5 ) {\fontsize{12}{12} \rotatebox{42} { \fontsize{12}{12} $L_{n-2} $}};
\node at (-0.3, 6 ) {\fontsize{12}{12} \rotatebox{-80} { \fontsize{12}{12} $L_{n-2} $}};
\end{tikzpicture}\caption{After breaking the CKC up the diagonal appears in both parts.}\label{fig:feasibleL}
\end{figure}
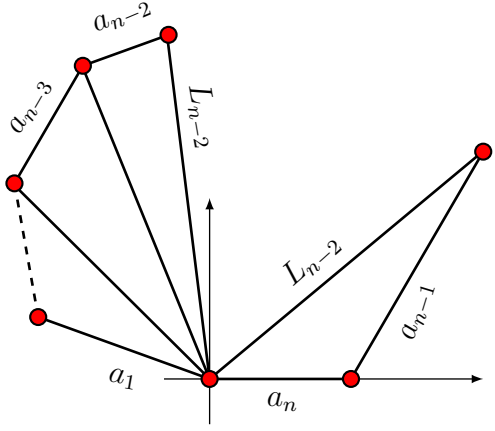

\begin{remark}[First step $k=1$]\label{rem:splitting} For simplicity, we assume in the following consideration that the links are ordered according to their length, i.e. $a_n \leq \dots \leq a_1$ applies. We have a closer look on the conditions for $L_{n-k-1}$ if $k=1$. Written in its expanded form we have
\begin{align*}
L_{n-2} \in  \Bigl[ a_n-a_{n-1}, a_n+a_{n-1} \Bigl]\cap\left[0 \vee  a_{n-2} - \sum_{j=1}^{n-3}a_j,\sum_{j=1}^{n-2}a_j \right].   
\end{align*}
This can be interpreted in the following way: Assume that we split the spherical configuration at the point connecting the links $a_{n-1}$ and $a_{n-2}$ and divide it into two parts in this way. The diagonal $L_{n-2}$ is the basis of the triangle with the remaining sides $a_n$ and $a_{n-1}$. Therefore satisfies $L_{n-2} \leq  a_n + a_{n-1}$, see Figure \ref{fig:feasibleL}. However, the length $L_{n-2}$ is also the base of the remaining part of the chain and must therefore also be a value that can be taken on by $||f_{a^n, n-2}(\al^{n-2}, \be^{n-2})||$. This implies that $L_{n-2} \leq \sum_{j=1}^{n-2} a_j $. Likewise, it is clear that $L_{n-2} \geq a_n-a_{n-1}$ and also, since the remaining chain must be able to reconnect with its first part $L_{n-2} \geq 0 \vee  a_{n-2}-\sum_{j=1}^{n-3}a_j $ has to hold. 
\end{remark}

By the definition of the space $\mathcal{DS}(a^n)$ the entries of its elements $L^{n-3}$ can be obtained recursively, resulting in a sampling strategy for the whole space. Theorem \ref{thm:reduction} ensures that the angles that can be obtained from $L^{n-3}$  satisfy the loop closure condition. We shortly summarize this in more detail in order to emphasize the progress of this work in comparison with the existing literature: 
\begin{itemize}
\item Sampling of the space $\mathcal{DS}(a^n)$ is an advance over its description by a system of linear inequalities as in \cite{han2006inverse,han2008conv}, since this requires the application of methods from linear programming. Furthermore, our theorem shows that $\mathcal{DS}(a^n)$ can be written as the intersection of a polytope and a cuboid. More precisely, if 
\begin{align}\label{eq:Pol}
\Pol(a^n):=\left\{(L_2,\dotsc,L_{n-2})\in \mathbb{R}^{n-3}_{\geq 0}   \colon L_{n-k-1} \in   \Bigl[\left|L_{n-k}-a_{n-k}\right|,L_{n-k}+a_{n-k}\Bigr]\right\},
\end{align}
denotes the polytope defined by nested intervals, then
\begin{align}\label{eq:Cub}
     \mathcal{DS}(a^n) = \Pol(a^n) \cap \Q(a^n), ~\text{with}~ \Q(a^n) :=  \prod_{1\leq k \leq n-3} \Bigl[0\vee \mathrm{    R}_{n-k-1}^{\min}, \mathrm{  R}_{n-k-1}^{\max}\Bigr].
\end{align}
\item Equation \eqref{eq:CGeqA} explicitly shows how joint angles of the CKC are related to its diagonals, which ultimately leads to a representation of the joint angles and thus also of $\Cs_{\text{alg}}(a^n)$ by the known set $\mathcal{DS}(a^n)$. Such an explicit relation  has been doubted in literature \cite[page 774]{JamesRTrinkleJ02b}. 
\end{itemize}
The description of $\mathcal{DS}(a^n)$ by the sampling procedure is already very useful. Nevertheless, we will further investigate $\mathcal{DS}(a^n)$  in section \ref{sec:diagspace}. For the moment, however, we would like to focus on the relationship of $\mathcal{DS}(a^n)$ with the work of \cite{han2007unified} and state one of its main result for convenience. 

\begin{theorem}[Feasible values for the diagonal lengths $L_{n-k}$, Section 2.3 of \cite{han2007unified}]
Let $\left(\al^{n-1},\be^{n-1}\right)\in M^{n-1} $  be a configuration of a KC. Then $\left(\al^{n-1},\be^{n-1}\right)$ is a spherical configuration if and only if its diagonals   $L_{1},\dots,L_{n-1} \in \mathbb{R}_{\geq 0}$ satisfy the following (in)equalities:
\begin{align}\label{eq:lihanineq}
\left\{
\begin{array}{rccllc} 
a_1 &=& L_1  &\,&   \, &      \\
\left|L_{k-1}-a_k\right| &\leq&  L_k  &\leq& L_{k-1}+a_k, & 2\leq k\leq n-1,\\  
a_k &\leq&  L_k + L_{k-1},  &   &     & 2\leq k\leq n-1,\\
a_n &=&   L_{n-1}  &   &     & 
\end{array} \right.
\end{align}
\end{theorem}
Since every configuration yields diagonal lengths that are contained in $\mathcal{DS}(a^n)$ and vice versa, we get the following result as a direct consequence of Theorem \ref{thm:reduction}.

\begin{corollary}[Equivalent system] Let $L_1 \dots, L_{n-1} \in \R_{\geq 0}$ with $L_1 = a_1 $ and $L_{n-1} = a_{n-1}$ satisfy the following system of inequalities:
\begin{align}\label{eq:ZanStein}
\left\{
\begin{array}{rccllc} 
\left|L_{k}-a_k\right| &\leq&  L_{k-1}  &\leq& L_{k}+a_k, & 3\leq k\leq n-1,\\  
0\vee\mathrm{    R}_{k}^{\min}  &\leq &  L_k &\leq    & \mathrm{    R}_{k}^{\max}     &  2\leq k\leq n-1
\end{array} \right.
\end{align}
Then any solution to system \eqref{eq:ZanStein} is also a solution to \eqref{eq:lihanineq}, which means that the system of inequalities are equivalent. 
\end{corollary} 

The following example illustrates the diagonal space for a simple CKCs and also confirms the statement in the corollary.

\begin{example}[Five bar mechanisms] We consider two CKCs with five links, also called five bar mechanisms. The first mechanism has link lengths that are all equal to one, i.e. $a^5 = (1,1,1,1,1)$. In this case we have 
\begin{align*}
\Pol\left( a^5 \right) = \left\{ (L_2, L_3)  \colon 0\leq L_3 \leq 2, \,  |L_3-1| \leq L_2 \leq L_3 +1  \right\} \, \text{and}\, \Q\left(a^5\right)= [ -1,3  ] \times [ 0, 2  ] \end{align*}
and $\mathcal{DS}\left(a^5\right)$ is the intersection of trapezoidal and a rectangle as depicted in Figure \ref{fig:DSpace}. The second mechanism has link lengths $b^5 = (2,3,4,2,3)$. For this mechanism we obtain 
$\mathcal{DS}(b^5)$ as the intersection of 
\begin{align*}
\Pol\left( b^5 \right) = \left\{ (L_2, L_3)  \colon 1\leq L_3 \leq 5, \,  |L_3-4| \leq L_2 \leq L_3 +4  \right\} \, \text{and}\, \Q\left(b^5\right)= [ 0,9  ] \times [ 1, 5  ]. 
\end{align*}
Consistently, also the inequalities in \cite{han2008conv} give the same domain.   
\end{example}

\begin{figure}
\begin{center}
\tdplotsetmaincoords{60}{110}
\pgfmathsetmacro{\rvec}{1.2}
\pgfmathsetmacro{\thetavec}{-30}
\pgfmathsetmacro{\phivec}{60}
\begin{tikzpicture}[scale=1.4,tdplot_main_coords]
\coordinate (O) at (0,0,0);
\draw[->] (0,0,0) -- (2,0,0) node[anchor=north east]{$x$};
\draw[->] (0,0,0) -- (0,2,0) node[anchor=north west]{$y$};
\draw[->] (0,0,0) -- (0,0,2) node[anchor=south]{$z$};
\tdplotsetcoord{P}{\rvec}{\thetavec}{\phivec}
\draw[color=black,thick] (O) --+(1,-1,1) --+(-1,-1,1)--+(0,0.4,1);
\draw[color=black,thick, dashed]  (0,0,0)--(0,0.5,1);
\draw[color=black,thick, dashed]  (0,0,0)--(-1,-1,1);
\fill[ball color=blue, opacity=0.15] (0,0,0) circle (0.9cm);
\fill[ball color=red, opacity=0.15] (-1,-1,1) circle (1.3cm);
\draw[line width = 1pt,red, rotate = -30] (-0.35,-0.6,0.08)   ellipse (1.15 and 0.23);
\draw[line width = 1.5pt,black, dashed,  rotate = -30] (-0.35,-0.6,0.08)   ellipse (1.15 and 0.23);
\node[rotate = -23] at (1, -0.5,0.5) {$a_1 = L_1$};
\node[rotate = 65] at (0, -1.2,1) {$a_2$};
\node[rotate = -25] at (0, -0.25,1.6) {$a_3$};
\node[rotate = 0] at (0, -0.7,1) {$L_2$};
\node[rotate = -20] at (0, 0.5,0.5) {$L_3$};
\shade[ball color=red!30!white] (0,0,0) circle (1.5pt);
\shade[ball color=red!30!white] (-1,-1,1) circle (1.5pt);
\shade[ball color=red!30!white] (1,-1,1) circle (1.5pt);
\shade[ball color=red!30!white] (0,0.45,1) circle (1.5pt);
\end{tikzpicture}\caption{A configuration of a CKC is constructed from a set of feasible diagonals: At the endpoint of the link $a_2$ a sphere of radius $a_3$ is centered and intersected by a sphere of radius $L_3$ centered on the origin. The link $a_3$ connects the endpoint of link $a_2$ with a point on the intersection, which is a circle. According to this construction the endpoint of $a_3$ hast distance $L_3$ from the origin. Our approach computes the parametrization of the intersection circles in terms of the joint angles of the CKC. }\label{fig:CKCconstruct} 
\end{center}
\end{figure}
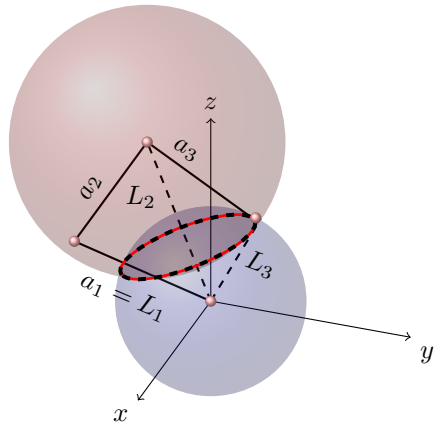

As soon as one obtains a feasible diagonal vector of length $L^{n-3}$, one can already geometrically construct a configuration of a CKC from it by intersecting spheres, see  Figure \ref{fig:CKCconstruct}. In the sequel, we determine the joint angles of the CKC from $L^{n-3}$. As expected, these parametrize the intersection circles which occur in the geometrical construction. This is numerically demonstrated in section \ref{sec:Appcon}.   

\begin{figure}
\centering
\begin{tikzpicture}[scale = 0.7]
\draw[line width = 0.5pt, gray] (0,0) grid (6,6);
\draw[->, -latex, line width = 0.5pt] (0,0) -- (7,0);
\draw[->, -latex, line width = 0.5pt] (0,0) -- (0,7);
\draw[line width = 1pt, fill = red, opacity = 0. 5] (0,2) -- (2, 0) -- (4, 2)--(4, 6 )--(0,2);
\draw[line width = 1pt, fill = red, opacity = 0. 5] (0,0) -- (6, 0) -- (6, 4)--(0, 4 )--(0,0);
\draw[line width = 1.5pt] (0,2) -- (2, 4) -- (4, 4)--(4, 2 )--(2,0)--(0,2);
\node at (6.5,0.5 ) {$L_3$}; 
\node at (0.5, 6.5) {$L_2$}; 
\node at (3.5, 4.6) {  \colorbox{red!50!white}{$\mathcal{P}$} }; 
\node at (5, 3.5) {  \colorbox{red!50!white}{$\mathcal{Q}$} }; 
\node at (2, -0.5) {  \colorbox{white}{$1$} }; 
\node at (-0.5, 2) {  \colorbox{white}{$1$} }; 
\node at (2, 2) {  \colorbox{red!75!white}{$\mathcal{P} \cap \mathcal{Q}$ } }; 
\end{tikzpicture}
\hspace{0.5cm}
\begin{tikzpicture}[scale = 0.5]
\draw[ line width = 0.5pt, gray] (0,0) grid (9,9);
\draw[->, -latex, line width = 0.5pt] (0,0) -- (10,0);
\draw[->, -latex, line width = 0.5pt] (0,0) --(0,10);
\draw[line width = 1pt, fill = red, opacity = 0. 5] (1,3) -- (4, 0) -- (5, 1)--(5, 9 )--(2,6)--(1,5)--(1,3);
\draw[line width = 1.5pt, ] (1,3) -- (3, 1)--(5,1) --(5, 5 )--(1,5)--(1,3);
\draw[line width = 1pt, fill = red, opacity = 0. 5] (0,1) -- (9, 1) -- (9, 5)--(0,5 )--(0,1);
\node at (9.5,0.5 ) {$L_3$}; 
\node at (0.5, 9.5) {$L_2$}; 
\node at (3.5, 5.6) {  \colorbox{red!50!white}{$\mathcal{P}$} }; 
\node at (7, 3.5) {  \colorbox{red!50!white}{$\mathcal{Q}$} }; 
\node at (1, -0.5) {  \colorbox{white}{$1$} }; 
\node at (-0.5, 1) {  \colorbox{white}{$1$} }; 
\node at (3, 3) {  \colorbox{red!75!white}{$\mathcal{P} \cap \mathcal{Q}$ } }; 
\end{tikzpicture}
\caption{Left: Right: The Domain $\mathcal{DS}(a^n)$ for a CKC with five links of length one as intersection of a trapezoidal domain (all configurations without loop closure constraint) with a rectangle. Right: The diagonal space of the CKC with links $(2,3,4,2,3)$. The same domain is shown in \cite{han2008stratified}.    }\label{fig:DSpace}
\end{figure}
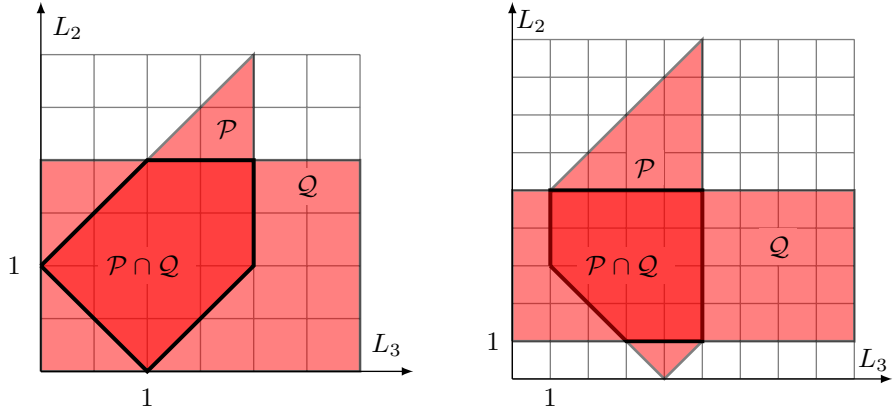

\begin{corollary}[Spherical configurations from diagonals]\label{thm:sampling} A spherical configuration $(\alpha^{n-1},\beta^{n-1}) \in \Cc \left( a^n \right)$, that is a configuration satisfying the equation \eqref{eq:closed_norm} can be obtained by the following procedure:\\ 
\begin{itemize}
\item[1.] Compute $R_k^{\min},R_k^{\max}$ for $2\leq k\leq n-1$. Then construct a vector with $n-3$ entries  \begin{align*}
L^{n-3} = \left( L_2, \dots, L_{n-2} \right) \in \mathcal{DS}({a^n})    
\end{align*}
of feasible diagonal lengths iteratively, starting from $L_1 := a_1$ and $L_{n-1} := a_n$ and continuing by choosing $L_{n-k-1}$ in $\left[|L_{n-k}-a_{n-k}|,L_{n-k+a}+a_{n-k}\right]\cap \left[0\vee R_{n-k-1}^{\min}, R_{n-k-1}^{\max} \right]$, for $k=1,\cdots,n-2$.\\  

\item[2.] For $2 \leq k \leq n-1$ assume  $(\al_{k-1},\be_{k-1})$ and the remaining angles $(\alpha^{k-2},\beta^{k-2})$ (for $k\geq 3$) are already related according to equation \eqref{eq:CGeqA} with $k$ replaced by $n-k+1$. Moreover, we can assume that $X_{k-1}, Y_{k-1}, Z_{k-1}$, $\Phi_{k-1}$ and $\Psi_{k-1}$ (when this is possible for $\Phi_{k-1}$ and $\Psi_{k-1}$) are already computed. Based on the angles already determined (including an arbitrary starting angle $(\alpha_1,\beta_1)$ in $[0,2\pi)\times[0,\pi]$), we now distinguish the following cases in the calculation of $\al_k$ and $\be_k$, which are in turn divided into subcases:
\begin{itemize}
\item[] \textbf{Return to the origin:}
$X_{k-1}^2  + Y_{k-1}^2+Z^2_{k-1}=0 \Leftrightarrow( X_{k-1}, Y_{k-1}, Z_{k-1})=0$. Then $(\alpha_k,\beta_k)$ can be arbitrarily chosen from $[0,2\pi)\times[0,\pi]$. If this occurs, the configuration of the CKC loops that return to the origin. Such cases have also been studied in  \cite[Section 2.2]{han2008conv}.\\%
Solutions: $(\alpha_k,\beta_k)\in [0,2\pi)\times[0,\pi]$.\\
\item[]  \textbf{Right angled triangle:}
This is the case when $L_{k-1}^2 + a_{k}^2 =  L_{k}^2$. If it happens, equation \eqref{eq:CGeqA}, with $k$ replaced by $n-k$, reduces to exactly this equation $L_{k-1}^2 + a_{k}^2 =  L_{k}^2$, which  means that $L_{k-1}, L_{k}$ and $a_k$ form a right-angled triangle (we exclude the already treated possibility of $X_{k-1}^2  + Y_{k-1}^2+Z^2_{k-1}=0$). It follows, that in this case 
\begin{align}\label{eq:compjointsA}
    \sin \left(\be_{k} + \Psi_{k} \right) \left[  \sin \left(\al_{k} +\Phi_{k-1} \right)^2  \left(   X_{k-1}^2  + Y_{k-1}^2 \right) + Z_{k-1}^2   \right]^{\frac{1}{2}}=0
\end{align}
(if all terms are defined) must vanish. This case leads to the following possibilities, all describing great circles on the unit sphere:\\

\begin{itemize}

\item $(X_{k-1}^2  + Y_{k-1}^2=0)\wedge Z^2_{k-1}\neq 0$. Then $L_{k-1}^2 =   Z_{k-1}^2 $. 
According to \eqref{eq:CGeqA0}, (replacing $k$ by $n-k$),  $\alpha_k$ is arbitrary in $[0,2\pi)$ and $\beta_k=\frac{\pi}{2}$. So the solution is the equator circle in the $(x,y)$-plane.\\ 
Solutions: $(\alpha_k,\beta_k)\in [0,2\pi)\times\left\{\frac{\pi}{2}\right\}$.\\
\item $(X_{k-1}^2  + Y_{k-1}^2\neq 0)\wedge (Z^2_{k-1}=0)\wedge\left(\sin(\alpha_k+\Phi_{k-1})^2=0\right)$. If $(X_{k-1}^2  + Y_{k-1}^2\neq 0)$ then it is possible to compute $\Phi_{k-1}$. Now $\sin(\alpha_k+\Phi_{k-1})^2=0$, meaning that $\alpha_k\in\{2\pi-\Phi_{k-1},\pi-\Phi_{k-1}\} (\!\!\!\mod 2\pi)$. Again $\beta_k$ is arbitrary in $[0,\pi]$.
The obtained solution set forms a longitudinal circle.\\
Solutions: $(\alpha_k,\beta_k)\in \left(\{2\pi-\Phi_{k-1},\pi-\Phi_{k-1}\}\times[0,\pi]\right).$\\
\item $ \sin \left(\al_{k} +\Phi_{k-1} \right)^2  \left(   X_{k-1}^2  + Y_{k-1}^2 \right) + Z_{k-1}^2 \neq 0$:  In this case  we choose $\alpha_k\in[0,2\pi)$ arbitrarily, compute $\Psi_k$ and since $\sin \left(\beta_{k} + \Psi_{k} \right)=0$ we have that
\begin{align*}
 \beta_k= \widetilde   \Psi_k:=\begin{cases}\pi-\Psi_k,\quad \Psi_k\in{[0,\pi)},\\ 2\pi-\Psi_k,\quad \Psi_k\in (\pi,2\pi]. \end{cases}   
\end{align*}
The solution yields a (skew) great circle on the unit sphere.\\
Solutions: $(\alpha_k,\beta_k)\in \left\{\left(\gamma,\widetilde{\psi}\right):\gamma\in[0,2\pi), \psi=\arg(\gamma+\Phi_{k-1},Z_{k-1})\right\}$.
\end{itemize}
\item[] \textbf{General triangle:} In this case we have to deal with  full extent of equation \eqref{eq:CGeqA0} or -- if $\Phi$ and $\Psi$ are defined -- equation \eqref{eq:CGeqA} (with $n-k$ replaced by $k$). We check again all appearing sub-cases.\\
\begin{itemize}
\item $(X_{k-1}^2  + Y_{k-1}^2=0)\wedge (Z^2_{k-1}\neq 0)$. In this situation we use equation \eqref{eq:CGeqA0} (replacing $n-k$ by $k$).  We conclude that  $\al_k \in [0, 2\pi )$ is arbitrary and obtain $\be_k$ by   
\begin{align*}
2a_{k}\cos(\beta_k) Z_{k-1}  &= L^2_{k}-a_k^2-L_{k-1}^2 \quad\Leftrightarrow\\
\beta_k  &=\arccos\left(\frac{L^2_{k}-a_k^2-L_{k-1}^2}{2a_k   Z_{k-1}}\right).
\end{align*}
The existence of the right hand side as a unique real number is guaranteed by the domain $[0,\pi]$ of $\beta_k$ and by $-1\leq  \tfrac{L^2_{k}-a_k^2-L_{k-1}^2}{2a_k Z_{k-1}}\leq 1$, since $Z_{k-1}^2=L_{k-1}^2$ and the lengths $L_k, a_k, L_{k-1}$ are the sides of a triangle. The solution set is a circle of latitude.\\
Solutions: $(\alpha_k,\beta_k)\in[0,2\pi)\times\left\{\arccos\left(\frac{L^2_{k}-a_k^2-L_{k-1}^2}{2a_kZ_{k-1}}\right)\right\}.$\\

\item In the case of $X_{k-1}^2  + Y_{k-1}^2\neq 0$ we infer that also $  \sin \left(\al_{k} +\Phi_{k-1} \right)^2  \left(   X_{k-1}^2  + Y_{k-1}^2 \right) + Z_{k-1}^2   \neq 0$, in particular $\sin \left(\al_{k} +\Phi_{k-1} \right)\neq 0$, otherwise we would be in the case $L_{k-1}^2+a_{k}^2 = L_{k}^2$ that has already been checked. Hence, $\Phi_{k-1}$ is defined. The value $\alpha_k$ can be determined by an inequality  analogue to \eqref{eq:CGineqsine} (replacing $n-1$ by $k$ and $n-2$ by $k-1$), giving 
\begin{align*}
\sin(\alpha_k+\Phi_{k-1})\in \left[-1,-\sqrt{ D_k \vee 0}\right]\cup\left[\sqrt{D_k \vee 0},1\right],~
\text{where} \quad D_k:=\frac{ \left(  L_{k}^2-a_{k}^2 -  L_{k-1}^2 \right)^2 -4a_{k}^2   Z_{k-1}^2  }{ 4 a_k^2(X_{k-1}^2 + Y_{k-1}^2)}.
\end{align*}
Taking the preimage of $\sin\colon[0,2\pi)\to[-1,1]$ of the above set and shifting it by $\Phi_{k-1} (\!\!\!\mod2\pi )$  yields the range for $\alpha_k$, in general the union of two intervals in $[0,2\pi)$. For this $\alpha_k$ compute now $\Psi_k$, which is defined in this case. The angle $\beta_k$ is then given by solving
\begin{align*}
\sin(\beta_k+\Psi_k)=\frac{L_{k}^2-a_k^2-L_{k-1}^2}{2a_k\left[  \sin \left(\al_{k} +\Phi_{k-1} \right)^2  \left(   X_{k-1}^2  + Y_{k-1}^2 \right) + Z_{k-1}^2   \right]^{\frac{1}{2}}},    
\end{align*}
which admits exactly one solutions for $\beta_k$ because the term on the right side equates to a number in $[-1,1]$ for the choice of diagonals and the precomputed quantities. The following set of solutions describe a (generic) circle on the unit ball.\\
Solutions: 
\begin{align*}
&\alpha_k\in\sin^{-1}\left(\left[-1,-\sqrt{D_k\vee 0}\right]\cup\left[\sqrt{D_k\vee0},1\right]\right)-\Phi_{k-1}\ \mod 2\pi,\\
\text{and\quad}&\beta_k = \sin^{-1} \left( \frac{L_{k}^2-a_k^2-L_{k-1}^2}{2a_k\left[  \sin \left(\al_{k} +\Phi_{k-1} \right)^2  \left(   X_{k-1}^2  + Y_{k-1}^2 \right) + Z_{k-1}^2   \right]^{\frac{1}{2}}} \right)-\Psi_k\ \mod 2\pi,\\
& \text{if this value } \beta_k\text{ is contained in }[0,\pi].
\end{align*}
\end{itemize}
\end{itemize}
\end{itemize}
\end{corollary}
With the corollary we can systematically find joint angles of a three-dimensional CKCs that satisfy the loop closure condition and have diagonal and it can therefore be used to sample the space $\Cc(a^n)$ effectively.\\     

At this point, we would like to deprive the reader of a direct consequence of our explicit calculations. As a by-product, we obtain an alternative proof of a well-known fact about the configuration space of three-dimensional CKCs.  

\begin{corollary} The configuration space of $\Cc(a^n) $ of a three-dimensional CKC is connected. 
\end{corollary}

\begin{proof} Theorem \ref{thm:sampling} shows that joint angles with the same diagonals are given by circles, when interpreted as points in the sphere $S^2$, compare with Figures \ref{fig:CKCconstruct} and \ref{fig:anglescorro}. Since $\mathcal{DS}(a^n)$ is connected, so is $\Cc(a^n)$. 
\end{proof}

\section{The diagonal space $\mathcal{DS}(a^n)$ in the case of three long links}\label{sec:diagspace}

In this section, we will further investigate $\mathcal{DS}(a^n)$. For a special class of CKCs we will be able to describe explicitly their diagonal lengths by a cube of dimension $n-3$. The class of CKCs  where this is true is related to the class of CKCs with {\em three long links}. We recall the definition in \cite[Section 3]{JamesRTrinkleJ02b}.

\begin{definition}[Long links] Let $L:= \sum_{i=1}^n a_i$ be the lengths of a CKC with links $a^n$. A subset $\mathcal{A}$ of its links $\mathcal{A} \subseteq \{a_1, \dots, a_n \}$ is referred to as containing long links of the CKC, iff the lengths of every pair of distinct links in $\mathcal{A}$ is strictly greater than $L/2$.     
\end{definition} 

Note that a CKC can have no more than three long links in which case the set $\mathcal{A}$ is unique. We can use a similar (weaker) notion to prove the following result for the polytopes $\Pol(a^n)$ and $\mathcal{Q}(a^n)$ from \eqref{eq:Pol}and \eqref{eq:Cub}.

\begin{theorem}\label{thm:longl} Assume a mechanism with link lengths $a_1\geq a_2\geq a_3\geq \dots \geq a_n $ be such that $\mathcal{A}=\{a_1, a_2, a_3\}$ and for all $a, b\in \mathcal{A}$ we have $a+b > \frac{L}{2}$. Then $\Pol(a^n) \subseteq \mathcal{Q}(a^n)$, and in particular, $\mathcal{DS}(a^n)=\Pol(a^n)$.
\end{theorem}

\begin{proof}
 For $\Pol(a^n)$ to be contained in $\mathcal{Q}(a^n)$, it is enough to show that all inequalities defining $\Pol(a_n)$
also satisfy the restrictions imposed by $\mathcal{Q}(a^n)$. Note that $L_{n-1}=a_n$ and consider
\begin{align}\label{eq:lihan2}
|L_{k+1}-a_{k+1}|\leq L_{k}\leq L_{k+1}+a_{k+1},\quad k=1,\dotsc,n-2.
\end{align}
We have to show that $$0\vee \mathrm{    R}^{\text{min}}_k\leq L_k\leq \mathrm{    R}^{\text{max}}_k, \,  k=2,\dotsc,n-1.$$

First note, that $\mathrm{    R}_2^{\text{min}}=a_1-a_2\geq 0$ and, as $a_1$ is the largest element, thus $$[0\vee \mathrm{R}_2^{\text{min}},\mathrm{    R}_2^{\text{max}}]=[a_1-a_2,a_1+a_2]$$ and for all $k\geq 3,$
$$[0\vee \mathrm{    R}_k^{\text{min}},\mathrm{    R}_k^{\text{max}}]=\left[0,\sum_{i=1}^k a_i\right],$$
since always two of the largest elements become subtracted when computing $\mathrm{    R}_k^{\text{min}}$. The lower bound is thus never a problem for $k\geq 3$. For the upper bound, the highest possible value for $L_k$ that may be reached by using the nested inequalities \eqref{eq:lihan2}, starting from $L_{n-1}=a_n$ and going only down to $k$, is $\sum_{i=k+1}^{n}a_i$. In this case, $\mathrm{    R}_k^{\text{max}}=\sum_{i=1}^k a_i>\sum_{i=k+1}^{n}a_i$ as the first sum always contains at least $a_1$ and $a_2$. 
Hence, $L_k\in [0\vee \mathrm{    R}_k^{\text{min}},\mathrm{    R}_k^{\text{max}}]$ is always satisfied for $k\geq 3$. It remains to check the range of $L_2$. For the upper bound, $L_2\leq \mathrm{    R}_2^{\text{max}}$ follows by the exactly same argument as for $k\geq 3$. In case of the lower bound, the nested inequalities yield $L_2\geq |L_3-a_3|$. By assumption, we know that $a_2+a_3\geq\frac{L}{2}$, i.e. $a_2+a_3\geq a_1+\sum_{i=4}^n a_i$. Equivalently, $a_3\geq\sum_{i=4}^n a_i+(a_1-a_2)$ and a fortiori, $a_3\geq\sum_{i=4}^n a_i$. Knowing from the upper bound argument before that $L_3\leq \sum_{i=4}^n a_i$, we have in any case $a_3>L_3$. Thus the absolute $|L_3-a_3|$ turns to $a_3-L_3$. But this means, $$|L_3-a_3|=a_3-L_3\geq a_1-a_2\quad \Leftrightarrow\quad a_3+a_2\geq a_1+L_3,$$ 
which is true since $L_3$ is bounded by $\sum_{i=4}^n a_i$. Therefore $L_2\geq |L_3-a_3|\geq a_1-a_2=R_2^{\text{max}}$. Thus all inequalities defining $\mathcal{Q}(a^n)$ are satisfied ($L_1$ is trivially set to $a_1$), concluding the proof.
\end{proof}

\begin{remark}[Generalization] We would like to point out that the statement of Theorem \ref{thm:longl} is still true when $a + b \geq  L/2$ holds. For ordered CKCs the result therefore holds for a slightly larger class of CKCs. 
\end{remark}

The configuration space of a CKC described in terms of its joint angles does not depend on the ordering of its links (up to homeomorphisms) \cite[Remark 1.2]{JamesRTrinkleJ02a}. However, the property exhibited in Theorem \ref{thm:longl} the diagonal space $\mathcal{DS}(a^n)$ does, as will be illustrated the in example below.


\begin{example}\label{ex:orderlinks}
We consider the diagonal spaces of the CKCs with the link lengths $a^5 = (6, 5, 4, 1, 1)$ and $\tilde a^6 = (4,1,6,5,1)$. In both cases the CKCs have three long links. In the ordered case the polytope $\mathcal{P}(a^6)$ is contained within the cuboid $a^6$. For the two CKCs we obtain
\begin{align*}
&\mathcal{DS}(a^6) \colon L_3 \in [0,2] \cap [0,15], \quad  L_2 \in [|L_3-4|,L_3+ 4 ] \cap [1,11] \\
&\mathcal{DS}(\tilde a^6) \colon L_3 \in [4,6] \cap [0,11], \quad  L_2 \in [|L_3-6|,L_3+ 6 ] \cap [3,5]
\end{align*}
The diagonal spaces are depicted in Figure \ref{fig:DSorder}.

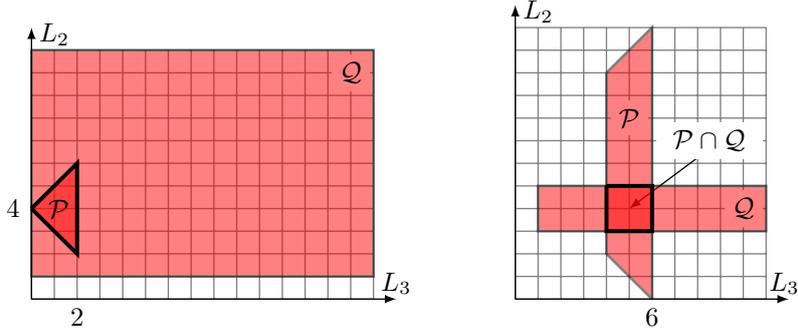
\begin{figure}
\centering
\begin{tikzpicture}[scale = 0.3]
\draw[line width = 0.5pt, gray] (0,0) grid (15,11);
\draw[->, -latex, line width = 0.5pt] (0,0) -- (16,0);
\draw[->, -latex, line width = 0.5pt] (0,0) -- (0,12);
\draw[line width = 1pt, fill = red, opacity = 0. 5] (0,1) rectangle (15, 11) ;
\draw[line width = 1pt, fill = red, opacity = 0. 5] (0,4) -- (2, 2) -- (2, 6)--(0, 4 );
\draw[line width = 1.5pt] (0,4) -- (2, 2) -- (2, 6)--(0, 4 );
\node at (16,0.7 ) {$L_3$}; 
\node at (1, 11.7) {$L_2$}; 
\node at (1.2, 4) {  
{$\mathcal{P}$} }; 
\node at (14, 10) {  \colorbox{red!50!white}{$\mathcal{Q}$} }; 
\node at (2, -0.8) {  \colorbox{white}{$2$} }; 
\node at (-0.8, 4) {  \colorbox{white}{$4$} }; 
\end{tikzpicture}
\hspace{1cm}
\begin{tikzpicture}[scale = 0.3]
\draw[line width = 0.5pt, gray] (0,0) grid (11,12);
\draw[->, -latex, line width = 0.5pt] (0,0) -- (12,0);
\draw[->, -latex, line width = 0.5pt] (0,0) -- (0,13);
\draw[->, -latex,line width = 0.5pt] (9,7) -- (5,4);
\draw[line width = 1pt, fill = red, opacity = 0. 5] (1,3) rectangle (11, 5) ;
\draw[line width = 1pt, fill = red, opacity = 0. 5] (4,10) -- (6, 12)--(6,0)--(4,2)--(4,10);
\draw[line width = 1.5pt] (4,5) -- (6, 5)--(6,3)--(4,3)--(4,5);
\node at (11.8,0.7 ) {$L_3$}; 
\node at (1, 12.7) {$L_2$}; 
\node at (5, 8) {  
 \colorbox{red!50!white}{$\mathcal{P}$} }; 
\node at (10, 4) {  \colorbox{red!50!white}{$\mathcal{Q}$} }; 
\node at (8.5, 7) {  \colorbox{white}{\fontsize{10}{10} $\mathcal{P}  \cap \mathcal{Q} $ } };
\node at (6, -0.8) {  \colorbox{white}{$6$} }; 
\end{tikzpicture}
\caption{Left: The diagonal space for the CKC with links $a^5 = (6,5,4,1,1)$. The polytope is clearly contained within $\mathcal{Q}(a^5)$. Right: The diagonal space for the CKC with links $\tilde a^5 = (4,1,6,5,1)$. Diagonal space consists of the square that is the  intersection of the trapezoidal and the rectangular domain. Interestingly the areas of both diagonal spaces are the same. Note that the dependence of the spaces on the link lengths is omitted in the notation in the pictures.}\label{fig:DSorder}
\end{figure}

\end{example}

Under the conditions of Theorem \ref{thm:longl}, it is certain that $\mathcal{P}(a^n)$ lies in $\mathcal{Q}(a^n)$. This result is so interesting for us because we are able to obtain a map, defined on a cube of dimension $n-3$,  whose image is the polytope $\mathcal{P}(a^n)$. A key ingredient to obtain this  parametrization is the following transformation of variables.

\begin{theorem}[Transformation of variables] Let $\Pol(a^n)$ be given as in \eqref{eq:Pol}. Define new variables $U_2, \dots, U_{n-2}$ according to the equation    
\begin{align}\label{eq:transform}
L_{n-k-1}^2 =  U_{n-k-1} + a_{n-k}^2   + L_{n - k}^2,  \quad 1 \leq k \leq n-3.  
\end{align}
Then, for $1 \leq k \leq n-3$, the $U_k$ satisfy the following system of inequalities:
\begin{align}\label{eq:transsys}
 | U_{n-k-1} |  \leq T_{n-k}(  U_{n-k}, \dots, U_{n-2}  ) := 2 a_{n-k} \sqrt{ \sum_{j = n-k}^{n-2} U_j + \sum_{j = n-k+1}^{n} a_{j}^2 }
 \end{align}
Note that for $k=1$  the value of the empty sum is zero and $T_{n-1}= 2a_na_{n-1} $.  
\end{theorem}

\begin{proof}
The inequalities defining $\Pol(a^n)$ are equivalent to  
\begin{align*}
\left(L_{n-k}-a_{n-k}\right)^2   \leq L_{n-k-1}^2  \leq   \left( L_{n-k}+a_{n-k} \right)^2,  
\end{align*}
 which is equivalent to 
\begin{align}\label{eq:transU}
    -2 L_{n-k} a_{n-k} \leq U_{n-k-1} \leq 2 L_{n-k} a_{n-k}  
\end{align}
For any $n$ it can be shown inductively that applying the substitution \eqref{eq:transform} repeatedly equation to \eqref{eq:transU} and thereby exploiting that $L_{n-1} = a_n$ yields system \eqref{eq:transsys}. 
\end{proof}

Analogously to Definition \ref{def:DiagSpace} we define the

\begin{definition} We define the transformed polytope space as the set 
\begin{align*}
\mathcal{PU}(a^n) = \left\{   U^{n-3} = (U_2, \dots, U_{n-2}) \in \R^{n-3} \colon ~\text{the entries of $ U^{n-3}$ satisfy \eqref{eq:transsys} } \right\}     
\end{align*}
\end{definition}

Obviously, the the space $\mathcal{PU}(a^n)$ can be transformed to $\mathcal{P}(a^n)$ by a transformation $F$ that can be obtained by using equation \eqref{eq:transform}. The next  example illustrates this transformation in more detail:   

\begin{example}[The space $\mathcal{PU}(a^n)$ for a five bar mechanism] We consider the mechanism with five members $a^n = (a_1, \dots, a_5)$ by which we want to illustrate the above theorem by performing the coordinate transformation explicitly. In this case the defining equations of $\Pol(a^n)$  in their squared form are given by  
\begin{align*}
 (a_5 - a_4)^2 \leq L_3^2  \leq  (a_5 + a_4)^2, \quad   (L_3 -a_3)^2 \leq L_2^2  \leq (L_3 +a_3)^2.  
\end{align*}
Expanding the squares and  using the substitutions $L_2^2 = U_2 + a_3^2 + L_3^2$ and $L_3^2 = U_3 + a_4^2 + a_5^2$ we obtain 
\begin{align*}
 -2a_5 a_4 \leq U_3  \leq 2a_5 a_4, \quad    -2 L_3 a_3   \leq U_2  \leq  2 L_3 a_3 , \end{align*}
where the right and left side for $U_2$ depend on $L_2$. In order to obtain inequalities in $U_2, U_3$ only, we express $L_3$ in terms of $U_3$ and obtain the inequalities for  
$\Pol(a^n)$ in $U$ space, that is 
\begin{align*}
-2 a_5 a_4 \leq U_3  \leq 2a_4 a_5, \quad    -2 a_3 \sqrt{U_3 + a_4^2 + a_5^2 }    \leq U_2  \leq  2  a_3 \sqrt{U_3 + a_4^2 + a_5^2 }. 
\end{align*}
\end{example}

In the case of CKCs with three long links we have by Theorem \ref{thm:longl} that  $\mathcal{DS}(a^n) = \mathcal{P}(a^n)$ and the polytope can be described by a new set of variables. More precisely, for CKCs with three long links we have that 
$F \left( \mathcal{PU}(a^n) \right) = \mathcal{DS}(a^n)$. The main contribution in this situation is that $ \mathcal{PU}(a^n)$ can be explicitly parameterized by a map $\Gamma$. This map is defined on a particularly simple domain, namely on a cube of dimension $n-3$. 

\begin{theorem}\label{thm:paramlong} For a CKC with that has three long links the space $ \mathcal{PU}(a^n)$ can described by parameters $s^{n-3} = (s_2, \dots ,s_{n-2})$ that are contained in the unit cube $I_{n-3} =  [-1,1]^{n-3}$. More precisely, the space $\mathcal{PU}(a^n)$ is given as the image of the map 
\begin{alignat*}{2}
\Gamma \colon I_{n-3} \rightarrow \mathcal{PU}(a^n), ~   s^{n-3} \mapsto   
            \left(
            \begin{array}{l}
            \Gamma_{n-2}( s_{n-2} )\\
            \Gamma_{n-3} (s_{n-3},s_{n-2})\\ 
            \vdots\\
            \Gamma_{2}(s_{2}, \dots, s_{n-2})
            \end{array}
            \right)
\end{alignat*}
where the components for $1 \leq k \leq n-3$ are recursively defined by 
\begin{align*}
\Gamma_{n-k-1} (s_{n-k-1},\dots,s_{n-2}) =  s_{n-k-1} T_{n-k}(\Gamma_{n-k}(s_{n-k},\dots,s_{n-2}) , \cdots, \Gamma_{n-2}(s_{n-2})   ).   
\end{align*}
\end{theorem}

\begin{proof} We have to show that the components of $\Gamma\left( s^{n-3} \right)$ satisfy system \eqref{eq:transsys} for $s^{n-3} = I_{n-3}$. For this purpose set   $U_{n-k} = \Gamma_{n-k}(s_{n-k},\dots, s_{n-2} )$  for $1 \leq k \leq n-3$. Then we have to show that 
\begin{align*}
 | U_{n-k-1} |  \leq T_{n-k}(  U_{n-k}, \dots, U_{n-2}  )   
\end{align*}
Plugging in the recursively defined entries of $\Gamma(s^{n-3})$  gives       
\begin{align*}
|\Gamma_{n-k-1}(s_{n-k-1},\dots, s_{n-2} )| &=  |s_{n-k-1}| \, T_{n-k}\left(  \Gamma_{n-k}(s_{n-k}, \dots, s_{n-2}),  \dots,  \Gamma_{n-2} (s_{n-2})  \right)\\   &\leq   
T_{n-k}(\Gamma_{n-k}(s_{n-k},\dots,s_{n-2}) , \cdots, \Gamma_{n-2}(s_{n-2})   ).
\end{align*}
One recognizes that both sides of the inequality contain the same term $T_{n-k}$.  The inequalities are therefore in any case fulfilled for $s_{n-k-1}\in [-1,1]$.
\end{proof}

\begin{remark} The description of $\mathcal{PU}(a^n)$ by parameters from $I_{n-3}$ is in general not a parametrization in the differential geometric sense, since the Jacobian of the map $\Gamma$ will not have full rank, if one of the terms $T_{n-k}$ vanishes. It is an interesting line of future research to investigate if this can happen in the case when a CKC has three long links. 
\end{remark}

We give an example that demonstrates the latter result:

\begin{example}[Six bar mechanism] Suppose a CKC has three long links and let its link lengths be given by $a^6$. In this situation the image of  the map $\Gamma \colon I_3 \rightarrow \mathcal{PU}(a^6)$ from Theorem \ref{thm:paramlong} is given by 
\begin{alignat*}{2}
\left(
\begin{array}{l}
\Gamma_4(s_4)\\
\Gamma_3(s_4, s_3)\\
\Gamma_2(s_4, s_3, s_2)
\end{array} 
\right)
=
\left(
\begin{array}{l}
s_4 T_{5}  \\
s_3  T_{4}(\Gamma_4(s_4) ) \\
s_2  T_{4}( \Gamma_3(s_3,s_4) ,\Gamma_4(s_4) )
\end{array} 
\right)
=
\left(
\begin{array}{l}
2 s_4 a_5 a_6  \\
2 s_3 a_4 \sqrt{ \Gamma_4(s_4) +a_5^2 +a_6^2}  \\
2s_2 a_3 \sqrt{ \Gamma_4(s_4) + \Gamma_3(s_3,s_4) +a_5^2 + a_4^2 + a_6^2  }
\end{array}
\right).
\end{alignat*}
Once the components of $\Gamma$ are computed, the map $F\colon \mathcal{PU}(a^6)  \rightarrow \mathcal{P}(a^6) =  \mathcal{DS}(a^6)$ according to \eqref{eq:transform} is given by 
\begin{alignat*}{2}
\left(
\begin{array}{l}
\Gamma_4(s_4)\\
\Gamma_3(s_4, s_3)\\
\Gamma_2(s_4, s_3, s_2)
\end{array} 
\right)
\mapsto
\left(
\begin{array}{l}
L_4(s_4)\\
L_3(s_3, s_4)\\
L_2(s_2,s_3, s_4)
\end{array}  
\right)
= 
\left(
\begin{array}{l}
\sqrt{ \Gamma_4(s_4) + a_{5}^2 +a_6^2 } \\
\sqrt{ \Gamma_3(s_4, s_3) + a_{4}^2 +L_4^2 }\\  
\sqrt{ \Gamma_2(s_4, s_3, s_2) + a_{3}^2 +L_3^2 }  \\
\end{array}  \right).
\end{alignat*}
\end{example}
\vspace{0.5cm}

\begin{remark}[Generalizability  of Theorem \ref{thm:paramlong} to unordered CKCs] Example \ref{ex:orderlinks} shows that the diagonal space depends on the order of its links. Also Theorem \ref{thm:paramlong} requires the links lengths to be given in descending order to guarantee that $\mathcal{P}(a^n) \subset \mathcal{Q}(a^n)$. This may seem restrictive at first, but we can easily generalize the result  of \ref{thm:paramlong} to CKCs with arbitrarily arranged links. For a permutation $\sigma \colon \{1, \dots, n\} \rightarrow \{1, \dots, n\}$ denote by $a = (a_1, \dots, a_n)$ and $a_\sigma = (a_{\sigma(1)}, \dots, a_{\sigma(n)})$ 
two vectors of link lengths that coincide up to the order entries and  let $\mathcal{DS}(a^n)$,   $\mathcal{DS}(a_\sigma^n)$ be their diagonal spaces. Then a one to one map$S \colon \mathcal{DS}(a^n) \rightarrow \mathcal{DS}(a_\sigma^n) $ between the diagonal spaces can easily be constructed in the following manner:   
\begin{itemize}
\item Uniquely assign angles to given feasible diagonals  $L^{n-3} \in \mathcal{DS}(a^n)$:
\begin{align*}
    L^{n-3} \mapsto (\al^{n-1}, \be^{n-1}) \in \Cc(a^n)
\end{align*}
\item Reorder the CKC in the joint angle space. To represent this rearrangement in terms of $\sigma$, we augment $(\al_{n}, \be_{n})$ to  $(\al^{n-1}, \be^{n-1})$, where $(\al_{n}, \be_{n})$ is chosen in such a manner that the link $a_n$, when attached to the end of link $a_{n-1}$,  reaches back to the origin. Then reordering in joint angle space results in 
\begin{align*}
(\al_1, \be_1, \al_2, \be_2, \dots, \al_{n}, \be_{n}) \mapsto (\al_\sigma^{n}, \be_\sigma^{n}) :=  (\al_{\sigma(1)}, \be_{\sigma(1)}, \al_{\sigma(2)}, \be_{\sigma(2)}, \dots, \al_{\sigma(n)}, \be_{\sigma(n)}). 
\end{align*}
Then $ || f_{a^n, n-1} (\al^{n-1},\be^{n-1} )|| = a_n $  and $ || f_{a_\sigma^n, n-1} (\al_\sigma^{n-1},\be_\sigma^{n-1} )|| = a_{\sigma(n)}$ and therefore $(\al_\sigma^{n-1},\be_\sigma^{n-1}) \in \Cc (a_\sigma^{n})$\\
\item Use the endpoint map for $(\al_\sigma^{n-1},\be_\sigma^{n-1}) $ to compute  $L_{\sigma, k} :=  || f_{a_\sigma^k, k} (\al_\sigma^{k},\be_\sigma^{k} )|| $  for $2 \leq k \leq n-2$, which gives $L_\sigma^{n-3}\in \mathcal{DS}(a_\sigma^n) $ an element in the diagonal space of the reordered chain.
\end{itemize}
The map $S$ is a bijection between diagonal spaces of CKCs with different link arrangements. Especially, if we have a parametrization for $\mathcal{DS}(a^n)$ under the assumptions of Theorem \ref{thm:paramlong}, this leads to a parametrization of  $\mathcal{DS}(a_\sigma^n)$ for an arbitrary arrangement. 
\end{remark}

\section{Numerical simulations}\label{sec:Appcon}

\begin{figure}
\centering
\includegraphics[width=1.0\textwidth]{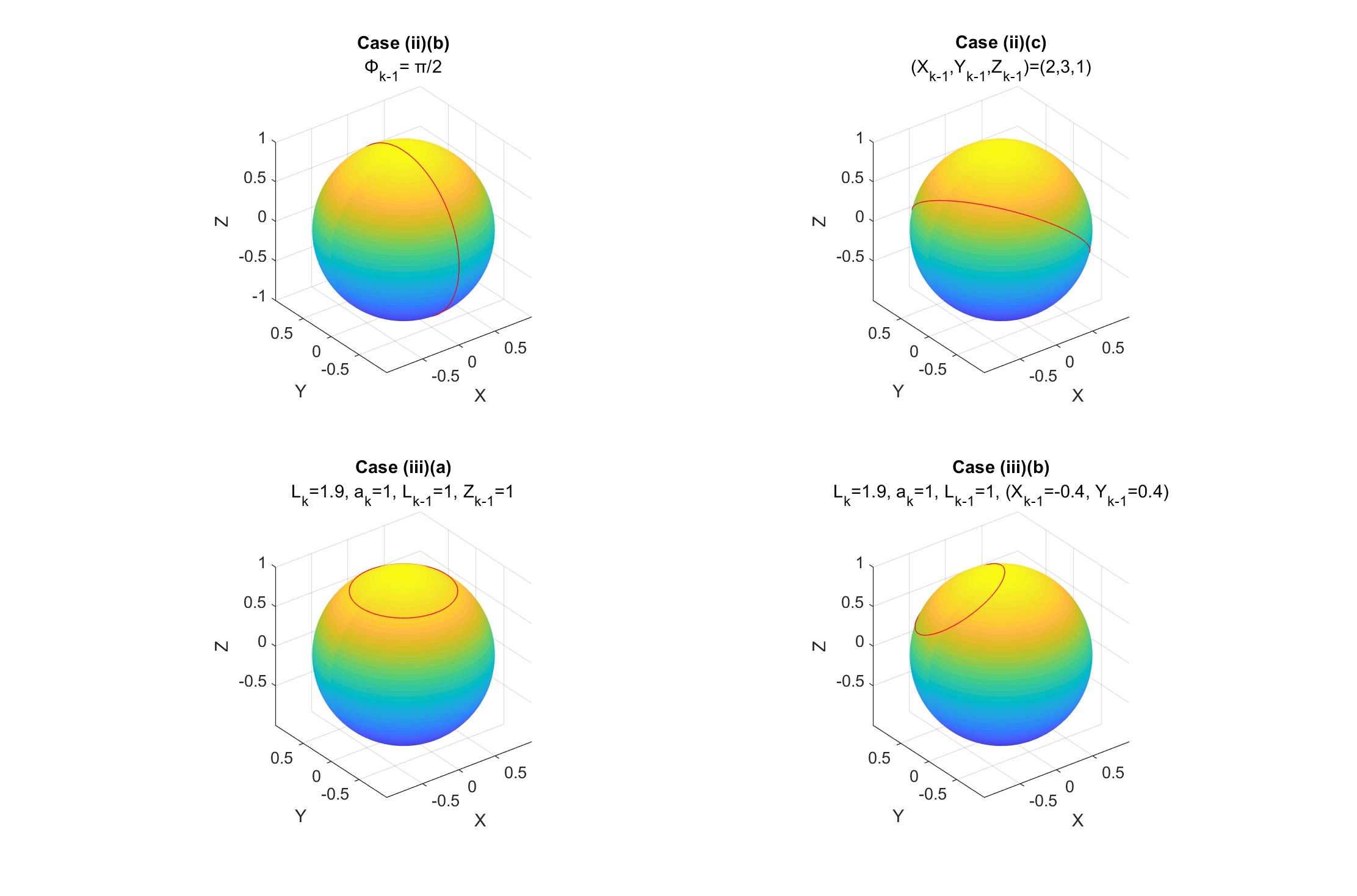}
\caption{The angles $\al_k,\be_k$ computed according to Corollary \ref{thm:sampling} for a certain choice of the variables already given. The angles are identified with their corresponding points on the unit sphere. As expected all possible solution values $(\al_k,\be_k)$ describe circles on the sphere. In the right angled case these circles are geodesic circles. Four different cases are depicted.}\label{fig:anglescorro}
\end{figure}

In this section we provide numerical examples that demonstrate the validity of the methods developed in this work. Firstly, we  illustrate the results from Corollary \ref{thm:sampling} that provides us with a procedure for obtaining the angles of a spherical joint when feasible diagonals are given. The solutions for diagonals corresponding to different solution cases are plotted, see Figure \ref{fig:anglescorro}.\\    

\begin{figure}
\centering
\includegraphics[width=1.0\textwidth]{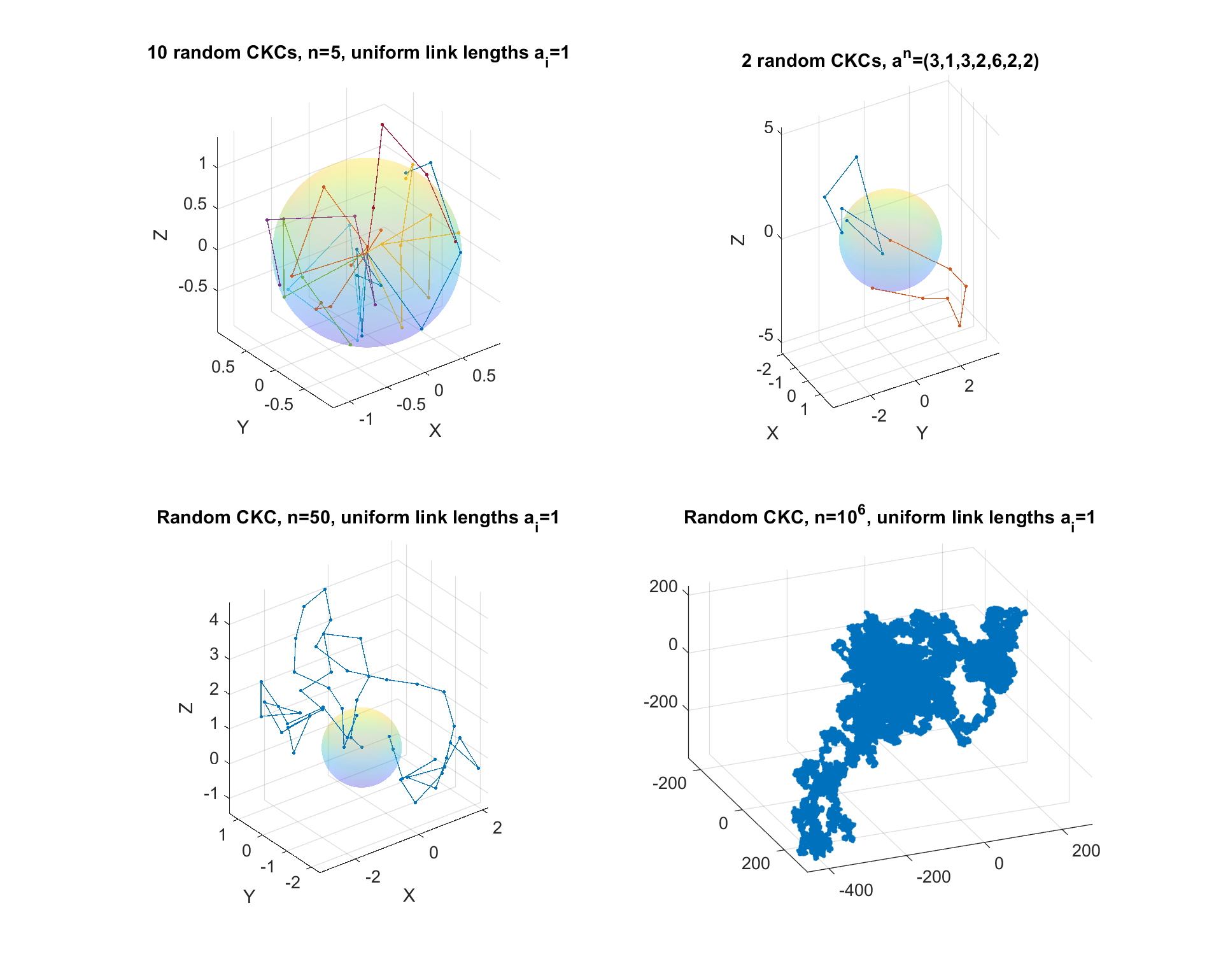}
\caption{Top left: Ten random configurations of CKCs with $n = 5$ links of length one are depicted. Note that the endpoint of each configuration lies on the sphere, which means that it can be connected with a link of length one to the origin. Top right: A single configuration of a CKC with different links is depicted. The last link reaches a sphere of radius two. Bottom left: A single configuration of a CKC with $n=50$ link of length one is depicted. The last link reaches a sphere of radius one. Bottom right: A configuration of a CKC with links all equal to one with one million links is depicted.  
}\label{fig:numconfigA}
\end{figure}

In our next numerical examples, we determine some random configurations of CKCs with $n \in  \{ 5, 7 , 50, 10^6 \}$ links. In the presented examples the following sampling strategy is used:
\begin{itemize}
    \item For $1\leq k \leq n-3$ we choose a sample uniformly at random of 
    \begin{align*}
        L_{n-k-1} \in \left[ | L_{n-k} -a_{n-k}|,  L_{n-k} + a_{n-k}\right].
    \end{align*}
    \item For the randomly chosen sample $L^{n-3}$ compute $D$ according to Theorem \ref{thm:sampling} and choose, uniformly at random again, 
    \begin{align*}
    \al_k \in \sin^{-1} \left( \left[ -1, -\sqrt{D_k \vee 0}\right]  \cup \left[ \sqrt{D_k \vee 0}, 1\right] \right) - \Phi_k.
    \end{align*}
     \item  Finally compute an angle $\be_k$ by the formula given  in Theorem \ref{thm:sampling}. 
\end{itemize}
We note that although in each step diagonals and angles are chosen uniformly at random, the steps are always interdependent. Also the distribution of the quantity $D_k$ will be different from the distribution of the chosen diagonals. However, since we only want depict configurations of CKCs based on samples we do not further discuss how they are chosen.  In Figure \ref{fig:numconfigA} several configurations are depicted and described in the caption. Our procedure is also feasible to systematically compute configurations also for very high dimensional CKCs. Figure \ref{fig:numconfigA} depicts a CKC with one million links. To the best knowledge of the authors such an example has not yet been presented in literature. The structure of this chain is reminiscent of a closed molecule consisting of a million atoms. Inspired by this examples we also speculate that our consideration of kinematic chains is relevant to the theory of closed continuous random walks (Brownian bridges).\\

\begin{figure}
\centering
\includegraphics[width=0.6\textwidth]{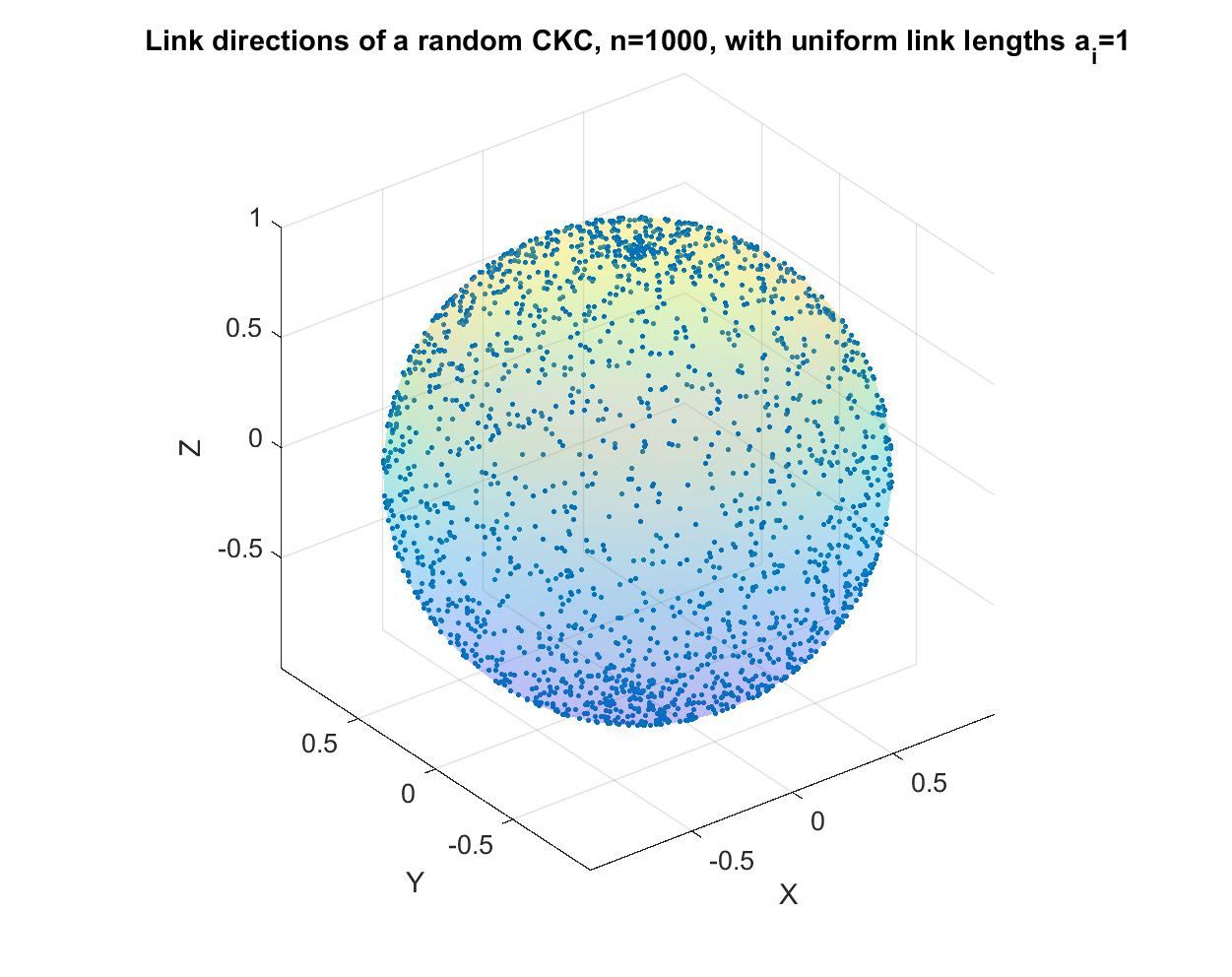}
\caption{A a random configuration of a CKC with $n=1000$ links of equal length one. The directions of the link segments of this configuration are shown as dots on the unit sphere. It is interesting to note that although directions are randomly chosen they are balanced due to the loop closure condition.}\label{fig:numconfigB}
\end{figure}

Configurations of CKCs with equal long links can also be represented graphically by thinking of all the rods of the CKC as being more attached at the origin. Since the rods are all the same length, we only draw their endpoints on a sphere with radius one. We find that this is an interesting way to represent a configuration, see Figure \ref{fig:numconfigB}.

\section{Conclusion and future work}
	
We have developed a new method for computing the diagonal lengths of a closed three-dimensional CKC.  From these, we can then determine the joint angles of the SJ of the CKC. Unlike existing work, our method does not require the solution of a system of linear inequalities by linear programming, nor does it rely on probabilistic methods. 
Numerical examples confirm the validity of the theoretical results of this work. We expect that our method will be useful for applications in motion planning for CKCs arising in robotics or protein kinematics.  An interesting line of future work would be a further investigation of the polytopes of the CKCs' diagonal space in more detail. Also a description of the diagonal space and the configuration space of a CKC with restrictions on the angles seems to be feasible with our method of directly tackling the closure equation. Since the sampling of configurations with random angles can be seen as a random walk with constrained probabilities, we believe that our work can also be useful in this context.


\bibliographystyle{plain}
\bibliography{CKCref.bib}


\end{document}